\RequirePackage{fix-cm}
\documentclass{svjour3}                     % onecolumn (standard format)
\smartqed  % flush right qed marks, e.g. at end of proof
\usepackage{graphicx}
\usepackage[colorlinks,citecolor=blue,urlcolor=blue,filecolor=blue,backref=page]{hyperref}
\usepackage{amsmath}
\usepackage[utf8]{inputenc} % allow utf-8 input
\usepackage[T1]{fontenc}    % use 8-bit T1 fonts
\usepackage{amssymb} 
\usepackage{wrapfig}
\usepackage{enumitem}
\usepackage{booktabs,etoolbox}
\usepackage{xcolor}
\usepackage{amsbsy}
\usepackage{mathtools}
\usepackage{subcaption}
\usepackage[ruled,linesnumbered]{algorithm2e}
\usepackage{url}            % simple URL typesetting
\usepackage{booktabs}       % professional-quality tables
\usepackage{amsfonts}       % blackboard math symbols
\usepackage{nicefrac}       % compact symbols for 1/2, etc.
\usepackage{natbib}
\usepackage{microtype}      % microtypography
\usepackage{epsfig}
\usepackage{epsf}
\usepackage{manfnt}
\usepackage{color}

\makeatletter
\newcommand{\oset}[3][0ex]{%
  \mathrel{\mathop{#3}\limits^{
    \vbox to#1{\kern-2\ex@
    \hbox{$\scriptstyle#2$}\vss}}}}
\makeatother

\newcommand{\ie}{\emph{i.e.}}
\newcommand{\eg}{\emph{e.g.}}

\newcounter{xxx}
\begin{document}

\title{Poisson Hyperplane Processes with Rectified Linear Units%\thanks{Grants or other notes
}
%\subtitle{SMSP}


\author{Shufei Ge         \and
        Shijia Wang \and Lloyd Elliott%etc.
}


\institute{S. Ge \at
              Institute of Mathematical Sciences, ShanghaiTech University, China\\
               \email{geshf@shanghaitech.edu.cn}           %\\
           \and
           S. Wang \at
              Institute of Mathematical Sciences, ShanghaiTech University, China\\
               \email{wangshj1@shanghaitech.edu.cn}               \and
           L. Elliott \at
              Department of Statistics and Actuarial Science, Simon Fraser University, Canada\\
               \email{lloyd\_elliott@sfu.ca}\\
}


\maketitle

\begin{abstract}
 Neural networks have shown
state-of-the-art performances in various classification and regression tasks. Rectified linear units (ReLU) are often 
used as activation functions for the hidden layers in a neural
network model. In this article, we establish the connection between the Poisson hyperplane processes (PHP) and  two-layer ReLU neural networks. We show that the PHP with a Gaussian prior is an alternative probabilistic representation to a two-layer ReLU neural network. In addition, we show that a two-layer neural network constructed by PHP is scalable to large-scale problems via the decomposition propositions. Finally,  we propose an annealed sequential Monte Carlo algorithm for Bayesian inference. Our numerical experiments demonstrate that our proposed method outperforms the classic two-layer ReLU neural network.  The implementation of our proposed model is available at \url{https://github.com/ShufeiGe/Pois_Relu.git}.
%\subclass{MSC code1 \and MSC code2 \and more}
\end{abstract}

\section{Introduction}\label{sec:intro}
Artificial neural networks are a popular set of machine learning methods, also called neural networks (NNs).  NNs were firstly introduced by \cite{mcculloch1943logical}, the early success of NNs sparked the first wave of interest in it during the 1960s and 1980s. %Other powerful machine learning methods with stronger mathematical foundations, such as support vector machines (SVMs), were invented around the 1990s and outperformed NNs, which put their development on hold \citep{geron2019hands}. 
Benefiting from the development of computing resources,  we are witnessing another wave of interest in NNs. As one of  the cores of deep learning and machine learning,  NNs have gained unprecedented attention for their outstanding performance in a variety of machine learning  or artificial intelligence tasks across different disciplines, especially with the recent successful launch of AlphaGo, AlphaFold and GPT~\citep{alphago,alphafold,Bahrini2023ChatGPTAO}.

Despite the great success of NNs in many applications, their black-box nature may lead to underestimation issues for problems that are not well tuned. In addition, training NNs is difficult as it involves a non-convex optimization problem, and the computational cost increases exponentially with the increase in the depth of the NNs.  Reducing the costs of  NNs remains a challenging task and a better understanding of how and why NNs work can lead to great improvements. This explains why recently NNs gain lots of interests in the theoretical studies. 
% A NN is defined by combinations of linear functions (\emph{neurons}) and non-linear  functions (\emph{activation functions}). Linear functions can be viewed as hyperplanes, researchers are trying to investigate the theories of NNs via hyperplanes and operators.

Attempts are being made to build a rigorous mathematical theory on the NNs. Some of them focus on investigating the theories of NNs via building connections  between NNs and existing machine learning methods, such as Gaussian process, mean field theory, decision trees and  optimization theories. \cite{williams1996computing} firstly showed that the limit of a single-layer fully-connected NN with an infinite number of neurons,  along with the prior over its parameters, converges to a  Gaussian process (GP). \cite{lee2017deep} further derived the exact equivalence between infinitely wide deep networks and GPs.  \cite{poole2016exponential} built a connection between Riemannian geometry with the  mean field theory and the signal propagation in generic NNs with random weights.   \cite{sethi1990entropy} showed that a decision tree could be restructured into a multi-layer NN by the systematic design of a class of layered  NNs.  
Conversely, \cite{aytekin2022neural} demonstrated that  feedforward  NNs with piece-wise linear activation functions can be represented by decision trees. \cite{balestriero2018mad} built a rigorous bridge between NNs and approximation theory via max-affine spline  operators.  Recently, \cite{chaudhari2018stochastic}
proved that stochastic gradient descent (SGD) performs variational inference under certain conditions and it converges to limit cycle for deep networks.

An NN is an arbitrary input-output mapping specified by combinations of linear operators (\emph{neurons}) and non-linear functions (\emph{activation functions}).
Activation functions of NNs usually are nonlinear functions applied to the output of nodes to avoid the entire NN degenerating into a single linear map. Common activation functions include the sigmoid activation function, the tanh function, the rectified linear unit (ReLU) activation function, the leaky ReLU function, the exponential linear unit (ELU), and so on. In this paper, we mainly consider the ReLU activation function, which only allows  a certain number of neurons to be activated and is much more computationally efficient than  activation functions such as the sigmoid and tanh functions. 

Generally, each neuron firstly processes the input $\boldsymbol{x}$ with a linear operator $\ell(\cdot)$, an activation function $\phi(\cdot)$ is then applied to the output of the linear operator $\ell(\boldsymbol{x})$, with output $\phi(\ell(\boldsymbol{x}))$.  Every linear operator $\ell(\cdot)$, in a multidimensional Euclidean space $\mathbb{R}^p$, can be viewed as a unique hyperplane $h$ in that space. Consequently, the input process of each neuron can be regarded as an activation function $\phi(\cdot)$ that processes the input $\boldsymbol{x}$ based on its position concerning the hyperplane $h$. Therefore, linear operators (\emph{neurons}) in the same layer of an NN could be viewed as a set of hyperplanes dividing the input space into multiple subspaces. The generative process of neurons in one-single layer of an NN could be viewed as a generative process of hyperplanes or a tessellation process in $\mathbb{R}^p$.   Inspired by this idea, we focus on constructing NNs via a tessellation process, for example, the Poisson hyperplane process (PHP) considered in this work. 

The two-layer ReLU NN is one of the popular NNs, it consists of an input layer, a hidden layer with the ReLU activation function, and an output layer. In this work, we investigate the relationship between the two-layer ReLU NN and the PHP. The PHP could be obtained by building its connection with the Poisson point process (PPP) based on the mapping theorem or the marking theorem \citep{last_penrose_2017_2}.  The PPP is well established and has been shown to possess elegant mathematical theorems, such as the {superposition theorem}, and the {restriction theorem}. Benefiting from these theorems, we  construct a two-layer ReLU NN by combining the PPP with a Gaussian prior, and further show that it is scalable to large-scale problems.

The main contribution of this work could be summarized with three points. Firstly, we investigate the relationship between the PHP and the two layer ReLU NN, and show that the PHP with a Gaussian prior is an alternative probabilistic generative model to the two-layer ReLU NN. 
 Secondly, we derive the decomposition propositions to justify that the two-layer ReLU NN constructed by the PHP is scalable to large-scale problems. Thirdly, we derive an annealed sequential Monte Carlo (SMC) algorithm for model inference.

The remainder of this article is organized as follows. We briefly review the two-layer ReLU NN and the PHP in Section \ref{sec:Preliminaries}.  Section \ref{sec: PHPtoNN} focuses on elaborating  the connection between the PHP and the two-layer ReLU NN.  In Section \ref{sec:DP}, we derive the decomposition propositions.   Model inference is given in Section \ref{sec: inference}.  Sections \ref{sec: sim} and \ref{sec: app} are numerical experiments.  We summarize our work and discuss the future work in Section \ref{sec:Dis}.

\section{Preliminaries}
\label{sec:Preliminaries}
In this section, we give a preliminary review of the two-layer ReLU NNs and the PHP.
\subsection{Basic Notations}
\label{sec:notation}
We use $n$ to denote the number of observations. We use the notation $\boldsymbol{1}(\cdot)$ for the $0-1$ valued indicator function.  We use bold lower-case letters for vectors, bold upper-case letters for matrices and plain lower-case letters for scalars, and $^\intercal$ for the transpose operation of matrices or vectors. We assume the input domain is $p$-dimensional and consider it in the Euclidean space. Denote $\boldsymbol{x}=(x_{1},\ldots,x_{p})^\intercal$, an arbitrary observation, where $x_{j}$ is the value of the $j$th coordinate, $j=1,\ldots,p$. Let $y$ be the response associated with $\boldsymbol{x}$. Denote $\mathbb{S}^{p-1}\equiv\{\boldsymbol{x}: ||\boldsymbol{x}||=1, \boldsymbol{x} \in \mathbb{R}^p \}$ the unit sphere in $p$-dimensional space, where $||\cdot||$ represents the norm, and let  $\overset{ }{\boldsymbol{n}}=(n_{1},\ldots,n_{p})^\intercal$ be a normal vector in $\mathbb{R}^{p}$.  A hyperplane $h$ can be defined uniquely by a pair of parameters $(\mu, \overset{ }{\boldsymbol{n}})$, $h\equiv \{\boldsymbol{x}\in \mathbb{R}^p: ~\langle\boldsymbol{x}, \overset{ }{\boldsymbol{n}}\rangle-\mu=0\}$, $\mu \in \mathbb{R}^+$, $\overset{ }{\boldsymbol{n}}\in \mathbb{S}^{p-1}$, where $\langle\cdot,\cdot\rangle$ represents the Euclidean scalar product. Similarly, we define $h^{-1}=(\mu, {\boldsymbol{n}})$.

\subsection{Two-layer ReLU Neural Networks}

A simple two-layer NN consists of an input layer, a hidden layer with activation functions, and an output layer. Activation functions utilize a pointwise  nonlinear function to the input to avoid the NN degenerating into a single linear map. In this paper, we mainly consider the rectified linear unit (ReLU) activation function,
$$\delta(c)=\max\{0,c\},~\forall ~c~\in ~\mathbb{R}.$$

Given an arbitrary input $\boldsymbol{x}\in \mathbb{R}^p$, a two-layer ReLU NN with $m$ neurons can be written in the following concise notation,
$$y=f\circ g^1\circ \delta \circ g^0 (\boldsymbol{x}),$$
where $\circ$ represents composition and $g^k: \mathbb{R}^{m_k}\longrightarrow \mathbb{R}^{m_{k+1}}$ maps values from $\mathbb{R}^{m_k}$ to $\mathbb{R}^{m_{k+1}}$, which are (vector) linear functions defined by  parameters $\boldsymbol{W}^{k}\in \mathbb{R}^{m_k\times m_{k+1}}$, $\boldsymbol{b^{k}}\in  \mathbb{R}^{m_{k+1}}$ ($k=0,1$), and  $f(\cdot)$ is usually set to be a simple identity function if the responses are continuous and  a (multinomial) logit link function if the responses are categorical.

As  previously described, $m_0=p,~m_1=m,~m_2=1$, $\boldsymbol{W}^0$ is a $p\times m$ matrix in which the $j$th ($j=1,\ldots,m$) column $\boldsymbol{W}^0_{\cdot j}$ represents the respective synaptic weights of the $j$th neuron, and $\boldsymbol{b}^0$ is an $m\times 1$ vector with the $j$th ($j=1,\ldots,m$)  element $\boldsymbol{b}^0_j$ representing the bias for the $j$th neuron. Similarly, $\boldsymbol{W}^1=(w_1,\ldots,w_m)^\intercal$ specifies the weights of the $m$ neurons with respect to the ReLU output,  and $\boldsymbol{b}^1=b^1$ is a scalar representing the bias.

Define $\boldsymbol{c}:=g^0(\boldsymbol{x};\boldsymbol{W}^0,\boldsymbol{b}^0)$, $\boldsymbol{z}:=\delta(\boldsymbol{c})$ and $\boldsymbol{o}:=g^1(\boldsymbol{z};\boldsymbol{W}^1,\boldsymbol{b}^1)$.
Generally, $g^0(\cdot)$ and $g^1(\cdot)$ take the following forms,
\begin{eqnarray*}
g^0(\boldsymbol{x};\boldsymbol{W}^0,\boldsymbol{b}^0)&=&\boldsymbol{b}^0+{\boldsymbol{W}^0}^\intercal\boldsymbol{x},\\
g^1(\boldsymbol{z}; \boldsymbol{W}^1,\boldsymbol{b}^1)&=&{b}^1+ {\boldsymbol{W}^1}^\intercal \boldsymbol{z},
\end{eqnarray*}
and 
\begin{eqnarray*}
    y &=& f(\boldsymbol{o}) +\epsilon,
\end{eqnarray*}
where $\epsilon$ represents the noise and usually is assumed to follow a normal distribution with zero mean.

\subsection{The Poisson Hyperplane Processes}\label{ppp}

The Poisson point processes (PPP) is well established, here we briefly introduce the generative of the Poisson hyperplane processes (PHP) by using its connection with the PPP.  %Fix a parameter $t \in  \mathbb{R}^+$, consider a stationary PPP on the real line with intensity $t$. Here $t$ is  the mean number of points per unit volume in $\mathbb{R}^p$. 
Denote by $\mathbf{P}(\mathbb{D})$ a point process defined on a measurable space $(\mathbb{D}, \mathcal{D})$, where $\mathbb{D}$ is a compact convex space in $\mathbb{R}^p$ and $\mathcal{D}$ is the Borel-$\sigma$ field of $\mathbb{D}$. For each $\mathbf{B}\subset \mathcal{D}$, denote $N_\mathbf{P}(\mathbf{B})$ the number of points of $\mathbf{P}$ within $\mathbf{B}$. The points count of $\mathbf{P}$ is Poisson distributed if for all $\mathbf{B}\subset \mathcal{D}$, $N_\mathbf{P}(\mathbf{B})\sim Poisson(\Lambda(\mathbf{B}))$,  where $\Lambda(\cdot)$ is a finite mean measure on $\mathbb{D}$. We say $\mathbf{P}$  is completely random if for all pairwise disjoint Borel sets $\mathbf{B}_1,\mathbf{B}_2,\ldots \subset \mathcal{D}$, $N_\mathbf{P}(\mathbf{B}_1)$, $N_\mathbf{P}(\mathbf{B}_2)$, $\ldots$ are independent of each other. A completely random point process with point counts being Poisson distributed is defined as a PPP \citep{last_penrose_2017}.   A stationary PPP on $\mathbb{D}$ with a finite mean $\Lambda(\mathbb{D})$ could be simulated by firstly sampling the number of data points $N_\mathbf{P}(\mathbb{D})$ from the Poisson distribution with mean $\Lambda({\mathbb{D}})$, and then generating $N_\mathbf{P}(\mathbb{D})$ points over the domain $\mathbb{D}$ according to the mean measure $\Lambda(\cdot)$.

For any point $(\mu,\overset{ }{\boldsymbol{n}}) $ defined on the Cartesian space  $\mathbb{R}^+\times \mathbb{S}^{p-1}\subset \mathbb{R}^p$,  there exists a map mapping each point uniquely to a hyperplane. We denote the map as $T(\cdot)$ and set $T((\mu,\boldsymbol{n}))=\{\boldsymbol{x}\in \mathbb{R}^p: \langle\boldsymbol{x},\overset{}{\boldsymbol{n}}\rangle-\mu=0\}$. Then, $T((\mu,\overset{ }{\boldsymbol{n}}))$ defines a hyperplane uniquely, where $\boldsymbol{n}$ is the unit normal vector of the plane and $\mu$ represents the signed distance from the origin to the plane. Conversely, for any hyperplane defined by a pair of parameters $(\mu,\overset{ }{\boldsymbol{n}}) \in \mathbb{R}^+\times \mathbb{S}^{p-1}$,  $T^{-1}=(\mu,\overset{ }{\boldsymbol{n}})$.

Note that the mapping $T(\cdot)$ described above is bijective for all hyperplanes (or points) defined by a pair of parameters $(\mu,\boldsymbol{n})$. Thus, for any compact convex domain 
$\mathbb{D} \subset \mathbb{R}^+\times \mathbb{S}^{p-1}$, there exists a measurable map $T(\cdot)$ that maps $\mathbb{D}$ to $\mathbb{H}$, $T: ~\mathbb{D} \rightarrow \mathbb{H}$, where $\mathbb{H}\equiv\{h:  T(h)^{-1} \in \mathbb{D}\}$ is a set of hyperplanes such that $T^{-1}(\mathbb{H})=\mathbb{D}$.  Therefore, any measure on $\mathbb{D}$ induces  a measure on $\mathbb{H}$, and vice verse.

A PPP in $\mathbb{R}^+\times \mathbb{S}^{p-1}$ could be obtained by leveraging the Marking Theorem in \cite{last_penrose_2017_2}, in which we associate each point $\mu$ of a PPP on $\mathbb{R}^+$ a unit normal vector $\boldsymbol{n}\in \mathbb{S}^{p-1}$ and in this case, a point  is specified by a pair of parameters $(\mu,\boldsymbol{n})$. To be specific, suppose $P(\mathbb{X})$ is a PPP in a measurable space $(\mathbb{X}, \mathcal{X})$ with finite intensity $\eta >0$, where $\mathbb{X} \subset \mathbb{R}^+$. Assign any $\mu \in \mathbb{X}$ a random mark $\boldsymbol{n}$ in a measurable space  $(\mathbb{S}^{p-1}, \mathcal{S}^{p-1})$, and denote $K$ to be the probability kernel from $\mathbb{X}$ to $\mathcal{S}^{p-1}$ such that $K(\mu, \cdot)$ is a probability measure for $\forall \mu \in \mathbb{X}$, and $K(\cdot, \boldsymbol{B})$ is measurable for each $\boldsymbol{B} \in \mathcal{S}^{p-1}$. By the Marking Theorem 5.6 of  \cite{last_penrose_2017_2}, the point process $P(\mathbb{X} \times \mathbb{S}^{p-1})$ is a PPP with intensity $\eta \otimes K$. In this article, we assume each point $\mu$ of a PPP on $\mathbb{R}^+$ is equally likely associated with a random unit normal vector $\boldsymbol{n}\in \mathbb{S}^{p-1}$, and $\boldsymbol{n}$ is uniformly distributed over $\mathbb{S}^{p-1}$, \ie~ $K(\mu, \cdot) \propto 1$ and $K(\cdot, \boldsymbol{B}) \propto \text{Volume}(\boldsymbol{B}) $.

As indicated in \cite{cltpp} and \cite{last_penrose_2017_2}, a PHP could be obtained by mapping a set of points of a {PPP} to a set of hyperplanes via $T(\cdot)$ defined above, which means that a PHP on $\mathbb{H}$ could be viewed as a PPP on $\mathbb{D}$ \citep{reitzner2013central,herold2021does}.

Denote  $\boldsymbol{H}$ the set of affine hyperplanes in $\mathbb{R}^p$,  $\mathcal{H}$ the Borel-$\sigma$ field of $\boldsymbol{H}$, and $\Psi(\cdot)$ a mean measure on $\mathcal{H}$. For any set $\boldsymbol{B}\in \mathcal{D}$, we write $\boldsymbol{H}_{\boldsymbol{B}} =\{h\in \boldsymbol{H}, T^{-1}(h)\in \boldsymbol{B}\}$. For any $\boldsymbol{B} \in \mathcal{D}$, we define  $\Psi(\boldsymbol{H}_{\boldsymbol{B}})=\Lambda(\boldsymbol{B})$. Assume the mean measure $\Psi(\cdot)$ is induced by the intensity measures $\psi()$, $\Psi(\boldsymbol{H}_{\boldsymbol{B}})=\int_{\boldsymbol{H}_{\boldsymbol{B}}}\psi(h)\mathrm{d}h$, and assume $\psi(\cdot)$ can be decomposed into Lebesgue measure on $\mathbb{R}^+$ and a uniform measure $\phi$ on $\mathbb{S}^{p-1}$, then

\begin{equation}
  \Lambda(\boldsymbol{B})=\Psi(\boldsymbol{H}_{\boldsymbol{B}})= \int_{\mathbb{S}^{p-1}}\int_{0}^{\infty} \boldsymbol{1}\left(T^{-1}((\mu,\boldsymbol{n}))\in \boldsymbol{B} \right) d\mu d\phi(\boldsymbol{n}).
  %t\Lambda(\cdot)=t\Psi(\cdot)= t\int_{\mathbb{S}^{p-1}}\int_{0}^{\infty} \boldsymbol{1}\left(T(\mu,\boldsymbol{n})\cap  \cdot \neq \varnothing  \right)  d\mu d\phi(\boldsymbol{n}).
  \label{lambdaM}
\end{equation}
The measure given in Eq.(\ref{lambdaM}) specifies a homogeneous, isotropic, and translation invariant PHP (or {PPP}).

\section{From Poisson hyperplane processes to two-layer ReLU neural networks}
\label{sec: PHPtoNN}

In this section, we build the connection between the PHP and the two-layer ReLU neural network.  
 Generally speaking,  a two-layer ReLU Bayesian neural network with Gaussian priors on the weights can be viewed as a PHP. The weights of PHP are assigned Gaussian priors. We shall elaborate it by considering a two-layer ReLU neural network with $m$ neurons in Euclidean space $\mathbb{R}^p$.

For an arbitrary $p$-dimensional input $\boldsymbol{x}$, the signal is firstly processed by a linear system $g^{0}(\boldsymbol{x};\boldsymbol{W}^0_{\cdot j},\boldsymbol{b}^0_{j})$ before being sent to neuron $j$ ($j=1,\ldots,m$) in the hidden layer. Neuron $j$ is activated and output $g^{0}(\boldsymbol{x};\boldsymbol{W}^0_{\cdot j},\boldsymbol{b}^0_{j})$ if $g^{0}(\boldsymbol{x};\boldsymbol{W}^0_{\cdot j},\boldsymbol{b}^0_{j})>0$, otherwise output $0$. The linear system is a linear combination of the $p$ predictors, which uniquely corresponds to a hyperplane $h_j=\{\boldsymbol{x}: g^{0}(\boldsymbol{x};\boldsymbol{W}^0_{\cdot j},\boldsymbol{b}^0_{j})=0,\boldsymbol{x} \in \mathbb{R}^p\}$ in $\mathbb{R}^p$, and the neuron will be activated if  observation $\boldsymbol{x}$ falls on the left side of the hyperplanes. Therefore,  the process from input to the hidden layer of a two-layer ReLU NN could be viewed as a generalization of a Poisson hyperplane process, the number of hyperplanes refers to the number of the neurons and each hyperplane corresponds to the signal process of each neuron. While each neuron may contribute differently to the output layer, which could be modelled by assigning weights from a Gaussian prior.

\begin{figure}[ht!]
     \centering
 \includegraphics[width=0.95\linewidth]{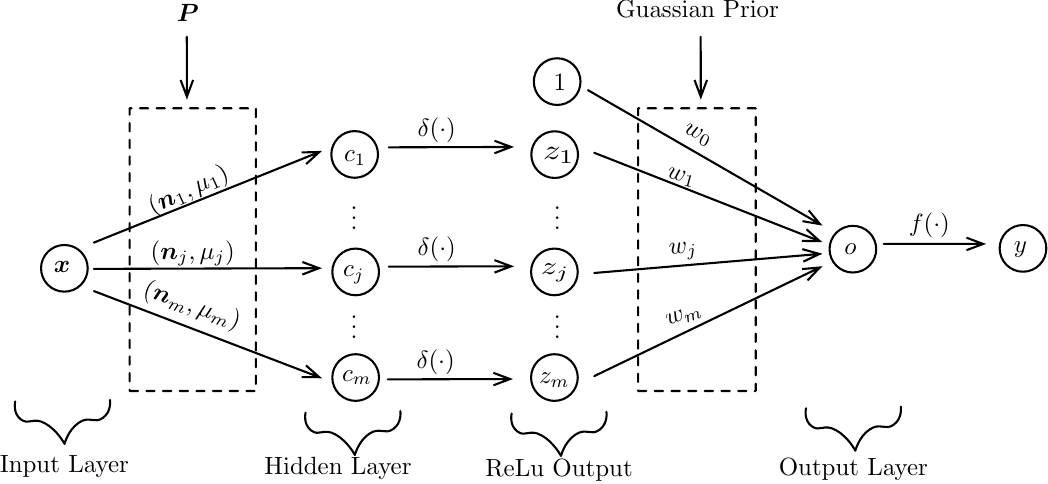}
     \caption{A PHP representation of a two-layer Bayesian ReLU neural network with Gaussian priors on weights.}
     \label{fig:relu2_pp}
 \end{figure}

To be more specific, recall that $\boldsymbol{P}(\mathbb{D})$ refers to a PHP concerning domain $\mathbb{D}$ with intensity $\Lambda(\cdot)$. Let $\boldsymbol{P}$ be a realization of the process, and  $|\boldsymbol{P}|$ represent the number of hyperplanes (or lines) of the realization. Therefore,  $\boldsymbol{P}$ can be parameterized by a set of hyperplanes (or lines) $h_j$, $\boldsymbol{P}=\{h_j, j=1,\ldots,|\boldsymbol{P}|\}$.  Each hyperplane $h_j$ can be represented by a  pair of a unit normal vector and the signed distance from the origin to the plane
as described in Section \ref{sec:notation}.
%represented by a set of points $\boldsymbol{x}$ specified by a pair of parameters $(\mu_j,\boldsymbol{n}_j)$, $h_j=\{\boldsymbol{x}: ~\langle\boldsymbol{x},~\overset{ }{\boldsymbol{n}_j}\rangle-\mu_j=0\}$,  $\overset{ }{\boldsymbol{n}_j}=(n_{j1},\ldots,n_{jp})^\intercal \in \mathbb{S}^{p-1}$ is a unit normal vector and $\mu_j\in \mathbb{R}^+$ indicates the signed distance from the origin to the plane.}
 Denote $\boldsymbol{z}$,  the ReLU output concerning data point $\boldsymbol{x}$ with a realization $\boldsymbol{P}$ of the PHP. Then
 $\boldsymbol{z}=(z_{0},z_{1},\ldots,z_{|\boldsymbol{P}|})^\intercal$, where ${z}_{0}=1$ is the constant term, $z_{j}=\delta(\langle\boldsymbol{x},~\overset{ }{\boldsymbol{n}_j}\rangle-\mu_j)$ indicates the result concerning $\boldsymbol{x}$ with respect to $j$th hyperplane of the PHP realization $\boldsymbol{P}$, $j=1,\ldots,|\boldsymbol{P}|$. 
 As indicated in the previous statement, $\boldsymbol{W}^0=(\overset{ }{\boldsymbol{n}_1}^\intercal,\ldots,\overset{ }{\boldsymbol{n}_m}^\intercal)$,  $\boldsymbol{b}_0=(\mu_1,\ldots,\mu_m)^\intercal$,  $\boldsymbol{z} = g^0(\boldsymbol{x};\boldsymbol{W}^0,\boldsymbol{b}_0)=\boldsymbol{x}\boldsymbol{W}_0-\boldsymbol{b}_0$, 
$\boldsymbol{W}^1=(w_1,\ldots,w_m)^\intercal$,  $\boldsymbol{b}_1=w_0$,  $\boldsymbol{o} = g^1(\boldsymbol{z};\boldsymbol{W}^1,\boldsymbol{b}_1)={w}_0+\sum_{j=1}^{m}w_jz_{j}$, $ y = f(\boldsymbol{o})+ \epsilon$, where $\epsilon$ represents the noise and usually is assumed to follow a normal distribution with zero mean. Figure \ref{fig:relu2_pp} illustrates a map from a Poisson hyperplane process with a Gaussian prior to a two-layer Bayesian ReLU neural network with Gaussian priors on weights.

%To be more precious, a PHP generates $m$ neurons by splitting the input domain with $m$ lines (\emph{or hyperplanes}), each line corresponds to a specific neuron in the two-layer neural network, and neurons may contribute differently to the output layer with weights sampled from a Gaussian prior. 

%The neuron will be activated if the observation falls on the left side of the line (\emph{or hyperplanes}). 

\section{Decomposition Propositions}
\label{sec:DP}
As described in Section~\ref{ppp}, the measurable function $T$ maps points set $\mathbb{D}$ in a one-to-one way onto the hyperplanes $\mathbb{H}$, therefore any PHP could be regarded as a PPP.   In this section, we propose three propositions to decompose the PPP into a finite set of independent PPP  by decomposing the intensity or the domain. With the help of these decomposition propositions, the PPP is scalable to large-scale problems, either large datasets or NNs with a large number of neurons.

Assume the mean measure $\Lambda(\cdot)$ is induced by the intensity measure $\lambda()$, \emph{i.e.}, $\Lambda({\boldsymbol{B}})=\int_{\boldsymbol{B}}\lambda(x)dx$.  As implied by the superposition theorem of \cite{last_penrose_2017} immediately, a PPP with intensity $\lambda$ could be decomposed into a finite number of independent PPP with intensity measure $\lambda_i$, such that $\sum_{i=1}^{K}\lambda_i=\lambda$, we refer to it as  \textbf{Proposition}~\ref{prop: 1}. As a complement to the restriction theorem of \cite{last_penrose_2017_2}, we have shown that the PPP possesses a consistency property as shown in \textbf{Proposition}~\ref{prop: 2}.  Combining  \textbf{Proposition}~\ref{prop: 1} and~\ref{prop: 2}, a PPP on domain $\mathbb{D}$ can also be decomposed into a finite set of independent PPP on pairwise disjoint subdomains such that the union of the subdomain is $\mathbb{D}$ as indicated in \textbf{Proposition}~\ref{prop: 3}. %, then aggregate the result of each properly. %According to \textbf{Proposition}~\ref{prop: 1}, a high-intensive PPP that requires expensive computer resources can be decomposed into multiple independent low-intensive PHP, each of which can be obtained at a relatively lower computer cost. Moreover, as a complement to the restriction theorem of \cite{last_penrose_2017_2}, we have shown that the PPP possesses a consistency property as shown in \textbf{Proposition}~\ref{prop: 2}.  Combining  \textbf{Proposition}~\ref{prop: 1} and~\ref{prop: 2}, a PPP on domain $\mathbb{D}$ can also be decomposed into a finite set of independent PPP on pairwise disjoint subdomains such that the union of the subdomain is $\mathbb{D}$ as indicated in \textbf{Proposition}~\ref{prop: 3}.  When the data scale is large, we could partition the data into multiple  groups by dividing the domain into a finite set of pairwise disjoint subdomains, and apply the PPP on each subdomain separately.%, then aggregate the result of each properly.

\begin{proposition}
\label{prop: 1}
If $\boldsymbol{P}_{\lambda}(\mathbb{D})$ is a PPP with $s-$finite intensity measure $\lambda$ on $(\mathbb{D},\mathcal{D})$, then it can be decomposed into a finite number, denoted as $K$, of independent PPP with $s-$finite intensity measure $\lambda_i$ on $\mathbb{D}$, such that $\sum_{i=1}^{K}\lambda_i=\lambda$.
\end{proposition}
\begin{proof}
Without loss of generality, we assume $\boldsymbol{P}_{\lambda_i}(\mathbb{D})$, $i=1,\ldots,K$, is a sequence of independent Poisson processes with $s-$finite intensity measure $\lambda_i$ on $(\mathbb{D},\mathcal{D})$ such that  $\sum_{i=1}^{K}\lambda_i=\lambda$.  
Write $\boldsymbol{P}':= \sum_{i=1}^{K}\boldsymbol{P}_{\lambda_i}(\mathbb{D})$, $\boldsymbol{P}'$ is a Poisson process with intensity measure $\lambda$ followed by the superposition theorem of \cite{last_penrose_2017}. 
Followed by Proposition 3.2 of \cite{last_penrose_2017}, for two Poisson processes $\boldsymbol{P}, \boldsymbol{P}'$ on  $\mathbb{D}$  with the same $s-$finite measure,   we have $$\boldsymbol{P'}\overset{D}{=}\boldsymbol{P}.$$ 
\end{proof}

%\begin{flushright}  
%\qedsymbol 
%\end{flushright}
\begin{proposition}
\label{prop: 2}
Let $\boldsymbol{P}_{\lambda}(\mathbb{D})$ be a PPP with $s-$finite intensity measure $\lambda$ on ($\mathbb{D},~\mathcal{D}$). For any $\mathbb{B}\in \mathcal{D}$, denote $\boldsymbol{P}_{\lambda}(\mathbb{D})\cap \mathbb{B}$ the induced process on $\mathbb{B}$ when  applying $\boldsymbol{P}_{\lambda}$ on $\mathbb{D}$ and then restricting it to subset $\mathbb{B}$. Then, $\boldsymbol{P}_{\lambda}(\mathbb{D})\cap \mathbb{B}$ is a PPP with $s-$finite intensity measure $\lambda$ on $(\mathbb{B},\mathcal{B})$, i.e. $\boldsymbol{P}_{\lambda}(\mathbb{D})\cap \mathbb{B}\overset{D}{=}\boldsymbol{P}_{\lambda}(\mathbb{B})$.
\end{proposition}
\begin{proof}
For any $\mathbb{C}\in \mathcal{B}$, $\mathbb{C}\subset \mathbb{B} \subset \mathbb{D} \in  \mathcal{D}$, we have $N_{\boldsymbol{P}\cap \mathbb{B}}(\mathbb{C})\sim Poisson(\lambda(\mathbb{C}) )$. For every $m\in \mathbb{N}_+$ and all pairwise disjoint sets $\mathbb{C}_1,\ldots,\mathbb{C}_m \in \mathcal{B}$, $\mathbb{C}_1,\ldots,\mathbb{C}_m \subset  \mathbb{B} \subset  \mathbb{D} \in  \mathcal{D}$, random variables  $N_{\boldsymbol{P}\cap \mathbb{B}}(\mathbb{C}_1)$, $\ldots$, $N_{\boldsymbol{P}\cap \mathbb{B}}(\mathbb{C}_m)$ are independent. Therefore, $\boldsymbol{P}_{\lambda}(\mathbb{D})\cap \mathbb{B}$ is a PPP on $(\mathbb{B},\mathcal{B})$ with intensity measure $\lambda$.
\end{proof}

\begin{proposition}
\label{prop: 3}
Let $\boldsymbol{P}_\lambda(\mathbb{D})$ be a PPP with $s-$finite intensity measure $\lambda$ on $(\mathbb{D},\mathcal{D})$, then it can be decomposed into a finite number, denoted as $K$, of independent PPP with intensity measure $\lambda$ on $\mathbb{D}_j$, i.e. $\boldsymbol{P}_\lambda(\mathbb{D})\overset{D}{=}\sum_{i=1}^{K}\boldsymbol{P}_\lambda(\mathbb{D}_i)$, where $\cup_{i=1}^{K} \mathbb{D}_i = \mathbb{D}$ and $\mathbb{D}_j\cap \mathbb{D}_i= \varnothing$ for $i\neq j$, $\mathbb{D}_i \in \mathcal{D}$,  $i, j=1,\ldots,K$.
\end{proposition}

\begin{proof}
Followed by \textbf{Proposition} \ref{prop: 2}, the sequence of independent PPP $\boldsymbol{P}_\lambda({\mathbb{D}_i})$, $i=1,\ldots,K$, is equivalently to a sequence of independent PPP with intensity measure $\lambda_i$ on $\mathbb{D}$, denoted as $\boldsymbol{P}'_{\lambda}(\mathbb{D})$, where $\lambda_i(\cdot)=\lambda \circ \boldsymbol{1}_{\mathbb{D}_i}(\cdot)$.  Denote $\boldsymbol{P}'=\sum_{i=1}^K\boldsymbol{P}'_{\lambda_i}(\mathbb{D})$, then $\boldsymbol{P}'$ is a PPP with intensity $\lambda'=\sum_{i=1}^{K}\lambda \circ \boldsymbol{1}_{\mathbb{D}_i}(\cdot)$ followed by \textbf{Proposition} \ref{prop: 1}.

Denote $L(\mu)$, $L'(\mu)$ the Laplace of functional of $\boldsymbol{P}_{\lambda}(\mathbb{D})$, and $\boldsymbol{P}'_{\lambda'}(\mathbb{D})$.  Let $\mathbb{R}_+(\mathbb{D})$ be all $\mathbb{R}_+$-valued measurable function on $\mathbb{D}$. For all $\mu \in \mathbb{R}_+(\mathbb{D})$,
\begin{equation*} \label{chf}
\begin{split}
L'(\mu) &=\exp\left[ -\int_\mathbb{D} (1- e^{\mu(x)}) \lambda'(dx) \right]\\
&=  \exp\left[ -\sum_{i=1}^{K}\int_\mathbb{D} (1- e^{\mu(x)}) \lambda'_i(dx) \right]\\
&=  \exp\left[ -\sum_{i=1}^{K}\int_{\mathbb{D}_i} (1- e^{\mu(x)}) \lambda(dx) \right]\\
&=  \exp\left[ -\int_{\mathbb{D}} (1- e^{\mu(x)}) \lambda(dx) \right]\\
&=L(\mu).
\end{split}
\end{equation*}

Followed by Theorem 3.9 of \cite{last_penrose_2017}, the Proposition has been proved as required.
\end{proof}

According to \textbf{Proposition}~\ref{prop: 1}, a high-intensity (\emph{i.e.} $\lambda$ is large) PPP that requires expensive computer resources can be decomposed into multiple independent low-intensity PHP, \emph{i.e.} decompose $\boldsymbol{P}_\lambda(\cdot)$ into a set of independent low-intensity PPP as long as $\sum_{i=1}^{K}\lambda_i=\lambda$.
Similarly, as indicated by \textbf{Proposition}~\ref{prop: 3}, when the data scale is large, we could partition the data into multiple groups by dividing the domain $\mathbb{D}$  into a finite set of pairwise disjoint subdomains $\mathbb{D}_i$, and apply the PPP on data points belonging to each subdomain separately.

Without using the decomposition propositions, we use high-intensity PPP to model the whole dataset, which includes more hyperplanes to split the domain. Hence, it requires more computer resources to finish the whole inference procedure. By using the decomposition properties, we can work on multiple independent low-intensity PPPs, or use independent PPPs to model disjoint sub-datasets. This is equivalent to using a simpler model (with a lower number of hyperplanes) or working on disjoint subsets of data. The inference procedure of each single job of the decomposition approaches is less intensive than the original inference.

\section{Model and Inference}
\label{sec: inference}

 In this section, we describe the probabilistic model of a two-layer ReLU NN given a PHP, and the inference scheme.  

\subsection{Model}
Assume $N$ pairs $(\boldsymbol{x}_i,y_i)$ input have been observed. As described in Section \ref{sec: PHPtoNN}, for input $\boldsymbol{x}_i$ ($i=1,\ldots, N$), neuron $j$ is activated if the observation $\boldsymbol{x}_i$ falls on the left side of the hyperplane $h_j$. Hence, the number of hyperplanes $|\boldsymbol{P}|$  equals the number of the neurons and each hyperplane corresponds to the signal process of each neuron.

Assume that the number of realized hyperplanes is prespecified (i.e. according to prior knowledge or tuning), conditioning on PHP $\boldsymbol{P}$ and weights of NN $\boldsymbol{w}$, the response is assumed to be normally distributed with mean $\langle\boldsymbol{w},\boldsymbol{z}^\intercal_{i \cdot}\rangle$, and variance $\sigma^2$.  The unknown parameters of interest includes PHP $\boldsymbol{P}$, weights of NN $\boldsymbol{w}$, variance parameter $\sigma^2$. We denote $\boldsymbol{\theta}=(\boldsymbol{P},\sigma^2,\boldsymbol{w})$, and let $p(\boldsymbol{\theta})$ denote the prior of $\boldsymbol{\theta}$. Our objective is to estimate the posterior distribution of $\boldsymbol{\theta}$. To benefit the inference, we assign conjugate priors for 
$\boldsymbol{w}$ and $\sigma^2$.
 %and $p(\boldsymbol{\theta})$ the prior of $\boldsymbol{\theta}$. 
The hierarchical structure of the model admits the following form:
    \begin{eqnarray*}
     \boldsymbol{P} &\sim&    \boldsymbol{P}(   \mathbb{D}),\\
     \sigma^2 &\sim& IG(a_0,b_0),\\
   %  \epsilon_i \sim N(0,\sigma^2),~i=1,\ldots,n\\
     w_j &\sim& N(\mu_0,\sigma_0^2), ~j=0,1,\ldots, |\boldsymbol{P}| ,   \\
     y_i|\boldsymbol{P},\sigma^2,\boldsymbol{w} &\sim& N(\langle\boldsymbol{w},\boldsymbol{z}^\intercal_{i \cdot}\rangle,  \sigma^2), i=1,\ldots,  N.  
\end{eqnarray*}
Here $IG(a,b)$ represents a inverse-gamma distribution with shape parameter $a$ and scale parameter $b$, $N(\mu,\sigma^2)$ represents a normal distribution with mean $\mu$ and variance $\sigma^2$, $\boldsymbol{P}(\mathbb{D})$ is a PHP defined on domain $\mathbb{D}$ with intensity $\Lambda= \int_{\mathbb{D}}\lambda(x)dx$.

The model performance-generalization ability trade-off could be achieved by tuning the number of hyperplanes. As we discussed in Section \ref{sec:Dis}, we could extend the model by treating the number of hyperplanes as an unknown parameter. In this case, the model bias-variance trade-off could be achieved
by tuning the hyper-parameters in the prior of PHP or adding a  hyper-prior distribution over the hyperparameters.

\subsection{Inference} 
For this hierarchical model, the posterior distributions of parameters $\sigma^2$ and $w_j$  are given in Eqs.(\ref{eq_par2}-\ref{eq_par1}),
\begin{eqnarray}
\label{eq_par2}
\sigma^2|\boldsymbol{P},\boldsymbol{w},\boldsymbol{y} &\sim& \text{IG}(a,b),\\
\label{eq_par1}
w_j|\boldsymbol{P},\sigma^2,\boldsymbol{y} &\sim& N(c_j,d_j), ~j=0,1,\ldots,\boldsymbol{|P|},
\end{eqnarray}
where $$a=a_0+\frac{N}{2}, ~b=b_0+\frac{1}{2}\sum_{i=1}^{N}\left(y_i-w_0-\sum_{j=0}^{\boldsymbol{|P|}}z_{ij}w_j\right)^2,$$
and $$c_j=\frac{\mu_0\sigma^2+\sigma^2_0\sum_{i=1}^{N}(y_i-w_0-\sum_{k\ne j}^{}z_{ik}w_k)z_{ij}}{\sigma^2+\sigma_0^2\sum_{i=1}^{N}z_{ij}^2},$$
$$d_j=\frac{\sigma^2\sigma^2_0}{\sigma^2+\sigma_0^2\sum_{i=1}^{N}z_{ij}^2}.$$

\begin{algorithm}
\caption{Metropolis-Hastings (MH) algorithm with random walk proposals}\label{MCMC1}
  {\bfseries Input:} a) Data $\boldsymbol{y},\boldsymbol{x}$; b) hyperparameters $a_0,~b_0$, $~\mu_0,~\sigma_0^2$; c) Target distribution $p(\boldsymbol{\theta}|\boldsymbol{y},\boldsymbol{x}) \propto p(\boldsymbol{y}|\boldsymbol{x},\boldsymbol{\theta})p(\boldsymbol{\theta})$; d) Total number of iterations $L$.  \\
 {\bfseries Output:} The MCMC samples of $\boldsymbol{P}$, $\sigma^2$ and $\boldsymbol{w}$.\\
 Set $t \leftarrow 0$, and initialize $\boldsymbol{P}$, $\sigma^2$, $\boldsymbol{w}$.\\
\While{$t \le  L$} {
 Set $t \leftarrow t + 1$.\\
 Update $\sigma^2$ according to Eq. (\ref{eq_par1}).\\
 Update $w_j,~j=1,\ldots,|\boldsymbol{P}|$ according to Eq. (\ref{eq_par2}).\\
Randomly sample $j$ from $\{1,\ldots,|\boldsymbol{P}|\}$, randomly sample $\overset{ }{n}_j\sim \frac{\Lambda(\cdot)}{\Lambda(S)}$, $\mu_j \sim  Uniform(0,1)$, denote  $h_j^*=\{\boldsymbol{x}: \mu_j+\langle\boldsymbol{x}, \overset{ }{n_j}\rangle=0\}$. Set $\boldsymbol{P}^*=\{\boldsymbol{P}/\{h_j\}\}\cup \{h_j^*\}$. Compute
$$\rho=min\left\{1,\frac{p(\boldsymbol{y}|\boldsymbol{P}^*,\sigma^2,\boldsymbol{w})}{p(\boldsymbol{y}|\boldsymbol{P},\sigma^2,\boldsymbol{w})}\right\}.$$\\
 Set $ \boldsymbol{P} \leftarrow \boldsymbol{P}^*$ with probability $\rho$.\\ 
} 
\end{algorithm}

We consider the case where the number of Poisson hyperplanes is fixed.  The posterior of the PHP $\boldsymbol{P}$ is intractable. One approach is to implement a Markov chain Monte Carlo (MCMC) algorithm to estimate the posterior. 
Assume the data domain is bounded by a ball $B_l\equiv \{\boldsymbol{x}:||\boldsymbol{x}||=l,\boldsymbol{x}\in \mathbb{R}^p\}$, $l\in \mathbb{R}^+$.
Algorithm \ref{MCMC1} displays the detail of PHP inference via a random walk Metropolis-Hasting (MH) algorithm. At every MCMC iteration, we first propose a new PHP. We sample $\mu_j \sim  \text{Uniform}(0, 1)$, and randomly generate a hyperplane by sampling a normal vector from the unit sphere and a signed distance from (0, $l$) uniformly. 
Then we compute the probability of accepting the new sample and decide whether to accept it or not by comparing with a uniform random number.

However, the discrete nature of the topology $\boldsymbol{P}$ induces a highly multimodal posterior space. The random walk MH algorithm  requires a large number of iterations to reach the stationary distribution due to the high rejection rate of updating $\boldsymbol{P}$. In this article, we develop a sequential Monte Carlo (SMC) algorithm for inferring $\boldsymbol{\theta}$ in the framework of annealing \citep{wang2018annealed}, to improve the efficiency.

There are several advantages of using SMC over standard MCMC algorithms. First, standard MCMC do not take advantage of highly parallel techniques straightforwardly, while the annealed SMC is an embarrassingly parallel method, in which a large number of particles can be run on different CPUs or GPUs simultaneously. Second, introducing a series of powered
posterior distributions can alleviate the issue of getting stuck in some local posterior region. Finally, convergence tests are required to make sure that MCMC methods are converged. SMC methods are built in the framework of importance sampling methods, hence the consistency property holds and the estimated posterior gets arbitrarily close to the true posterior when the number of particles goes to infinity \citep{chopin2004central}.

\begin{algorithm}
\caption{An annealed sequential Monte Carlo for inference }\label{asmc}
  {\bfseries Input:} a) Data $\boldsymbol{y},\boldsymbol{x}$; b) hyperparameters $a_0,~b_0$, $~\mu_0,~\sigma_0^2$;
  c) prior over $\boldsymbol{\theta}$;
d) likelihood function $p(\boldsymbol{y}|\boldsymbol{\theta},\boldsymbol{x})$;
e) a sequence of annealing power terms $\phi_0=0<\phi_1<\ldots <\phi_R=1$;
f) total number of particles $L$.  \\
 {\bfseries Output:} Approximation of the posterior distribution $\sum_{t=1}^{L}\frac{{w}_{R,t}}{\sum_{k=1}^{N}{w}_{R,k}}\boldsymbol{1}(\Tilde{\boldsymbol{\theta}}_{R,t})$\\
 Initialize SMC iteration index $r \leftarrow 0$. \\
  Initialize annealing power $\phi_r  \leftarrow 0$. \\
  %\For{\texttt{<some condition>}}
 %\State \texttt{<do stuff>}
%\EndFor
    \For{$t \gets 1$ to $L$} {
      Initialize particles $\boldsymbol{\theta}_{0,t}  \leftarrow \boldsymbol{\theta}\sim p(\cdot)$.\\
        Initialize weights to unity: $w_{0,t}  \leftarrow 1$.
        }
            \For{$r \gets 1$ to $R$} { 
            \For{$t \gets 1$ to $L$}{
            Sample particles $\Tilde{\boldsymbol{\theta}}_{r,t}\sim K_r(\boldsymbol{\theta}_{r-1,t},\cdot)$,  where $K_r$ is a $\pi_r$-invariant MH kernel (\ie~ one MH step described in Algorithm \ref{MCMC1}).\\
            Compute unnormalized weights:  $w_{r,t}=p(y|\Tilde{\boldsymbol{\theta}}_{r,t}, \boldsymbol{x})^{\phi_r-\phi_{r-1}}$.
            }
            \If{$r<R$}{
            Resample  $\boldsymbol{\theta}_{r,t}$ from $\Tilde{\boldsymbol{\theta}}_{r,t}$ with probability proportional to $w_{r,t}$,  $t=1,\ldots, L$.
            }
        }
        Return the weighted particle samples $({w}_{R,t},\Tilde {\boldsymbol{\theta}}_{R,t})$, $t=1,2\ldots, L$.
\end{algorithm}

We first define a sequence of distributions $\gamma_r(\boldsymbol{\theta})=p(\boldsymbol{y}|\boldsymbol{\theta})^{\phi_r}p(\boldsymbol{\theta})$ to facilitate the exploration of posterior surface. Here $\phi_r$ $(r = 1, 2, \ldots, R)$ is a power term, $\phi_0=0<\phi_1<\ldots<\phi_R=1$.  A small value of $\phi_r$ flattens the posterior distribution, and hence facilitate the exploration of the target. 
The intermediate target distribution $\pi_r(\boldsymbol{\theta})$ of the annealed SMC takes the form $\pi_r(\boldsymbol{\theta})\propto \gamma_r(\boldsymbol{\theta}) = p(\boldsymbol{y}|\boldsymbol{\theta})^{\phi_r}p(\boldsymbol{\theta})$. At each SMC iteration, we iterate between the following three  steps to propagate samples to approximate the next intermediate target distribution $\pi_r(\boldsymbol{\theta})$: {\it propagation}, {\it weighting} and {\it resampling}. Algorithm \ref{asmc} summarizes the annealed SMC. The {\it propagation} step is achieved by running a $\pi_r$-invariant MH kernel $K_r$, a single step of the MH algorithm described in Algorithm \ref{MCMC1}). For each propagated sample $\Tilde{\boldsymbol{\theta}}_{r,t}$ $(t = 1, 2, \ldots, L)$,  we evaluate the unnormalized weights according to  $w_{r,t}=p(y|\Tilde{\boldsymbol{\theta}}_{r,t}, \boldsymbol{x})^{\phi_r-\phi_{r-1}}$. The resampled particles are denoted by $\boldsymbol{\theta}$.
Finally, we conduct a resampling step to prune particles with small weights. After running the annealed SMC, we obtain a list of weighted samples $({w}_{R,t}, \Tilde{\boldsymbol{\theta}}_{R,t})$$(t=1, 2, \ldots, L)$ to represent the posterior of $\boldsymbol{\theta}$.

When the number of dimension is fixed, the computational complexity of Algorithm \ref{asmc} is $\mathcal{O} (NRL|\boldsymbol{P}|)$, which is a linear function of the number of observations $N$, the length of annealing power series $R$, the number of hyperplanes $|\boldsymbol{P}|$, and the number of particles $L$. By \textbf{Proposition}~\ref{prop: 1}, if we decompose a PPP with a large number of realizations of hyperplanes into $K$ independent PPP, each with $|\boldsymbol{P}|/K$ realizations of hyperplanes, the computational complexity of each PPP will be reduced to  $\mathcal{O} (\frac{NRL|\boldsymbol{P}|}{K})$. Similarly,  if we divide a large-scale dataset into  multiple disjoint data by domain, assume that the maximum of sub-dataset sizes is $m$ and apply the PPP to each sub-dataset, the computational complexity for each will be reduced up to $\mathcal{O} ({mRL|\boldsymbol{P}|})$. We also demonstrate the computational efficiency of the decomposition approaches via simulation studies in Section~\ref{sec: sim}.

\section{Simulation}
\label{sec: sim}
\begin{figure}[ht!]
     \centering
 \includegraphics[width=1.0\linewidth]{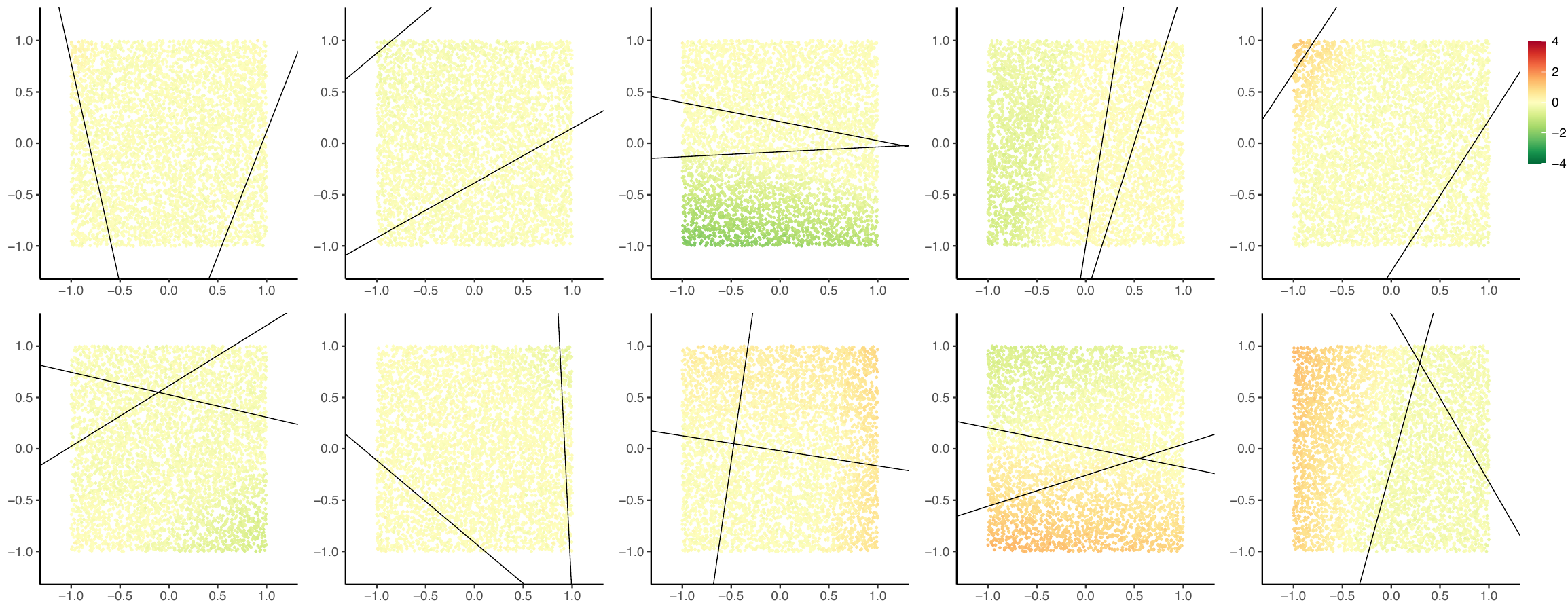}
     \caption{Visualization of $10$ randomly selected simulated datasets in the first simulation study. The horizontal-axis and vertical-axis are values of dots' coordinates. Dot color indicates the value of response $y$, and the black lines represent the corresponding generated lines.}
     \label{fig:sim}
 \end{figure}

In this section, we demonstrate the effectiveness of our proposed method (\emph{pois}) on a set of simulated data. 
 %, by comparing with decision trees (\emph{dt}), random forests (\emph{rf}), support vector machines with radial kernels (\emph{svm}$\_r$), support vector machines with linear kernels (\emph{svm}$\_l$) and  linear regression model (\emph{lm}).  
 In our first simulated study, each dataset is simulated by randomly drawing $5,000$ paired coordinates $(x_1,x_2)$ from domain $[-1,-1]\times [-1,-1]$. %, and points fall out of circle $x_1^2+x_2^2=2$ are removed. 
Two random lines $l_1(x_1,x_2)=0$, $l_2(x_1,x_2)=0$ intersecting the plate are randomly generated to separate data points into different groups. The response values are given by $y=w_0+w_1\delta(l_1(x_1,x_2))+w_2\delta(l_2(x_1,x_2))+\epsilon$, where $w_i$ $(i=1, 2)$ is drawn from the standard normal distribution, $w_0$ and the noise $\epsilon$ is sampled from $N(0, 0.1^2)$. $100$ simulated data are generated and Figure \ref{fig:sim} provides a visualization of $10$ randomly simulated datasets. 

\begin{figure}[ht!]
     \centering
  \includegraphics[width=1\linewidth]
 {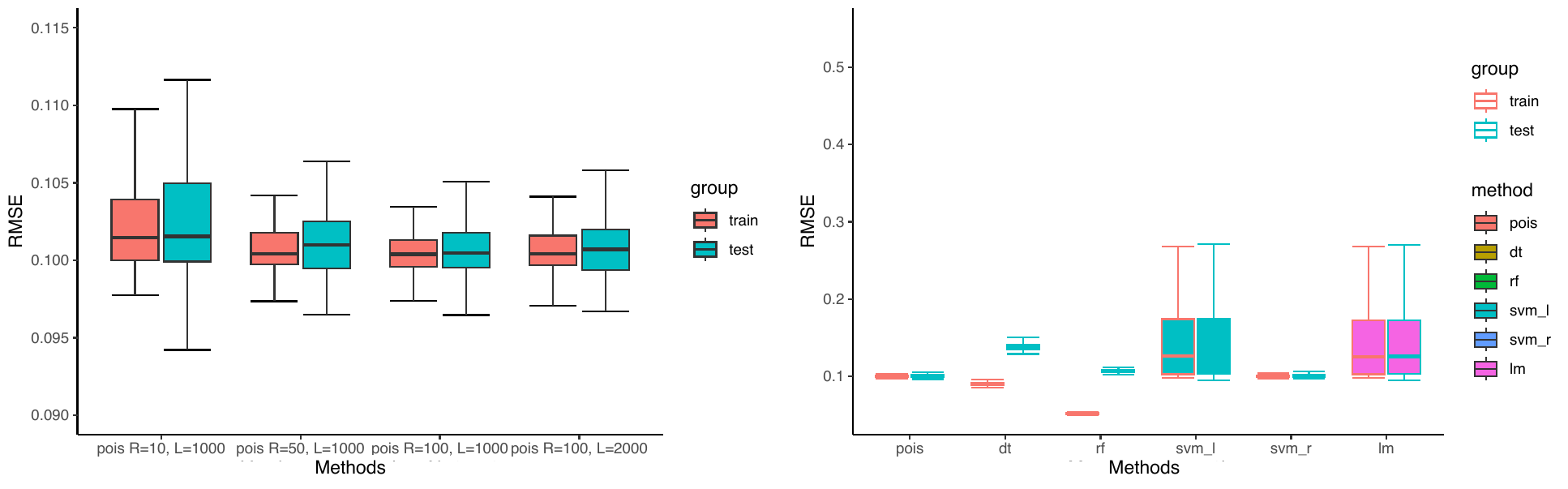}
     \caption{RMSE across different methods on train and test data splits  generated in the first simulation study. The solid boxes (lines) represent the RMSE on the train datasets and the dotted ones indicate the values with respect to the test datasets.  \emph{Left)} The RMSE of our proposed method with varying $L$ and $R$.  \emph{Right)} Classical machine learning methods and our proposed model ($L=1000,~R=100$). The  number of hyperplanes of our proposed methods is fixed to $2$ in these experiments. 
     }
     \label{fig:relu2}
 \end{figure}

We conduct experiments to investigate the impact of the number of particles and the length of power sequences on model inference. For each dataset, we randomly generate  train/test splits, with each train and test data containing 75\% and 25\% of the samples. We consider the following scenarios:  $R=10, 50, 100$ with $L = 1000$ and $L=2000$ with $R = 100$.  As shown in Figure \ref{fig:relu2} (\emph{Left}), larger $L$ and $R$ values tend to lead to smaller RMSE on test data set. However, the computational cost  increases linearly with $R$ (or $L$). In order to balance the computational cost, in the following experiment, if there is no further description, we set $L=1000,~R=100$. We also compare the performance of our proposed method (\emph{pois}) with classical machine learning methods on simulated data, including decision trees (\emph{dt}), random forests (\emph{rf}), support vector machines with linear kernels (\emph{svm}$\_l$), support vector machines with radial kernels (\emph{svm}$\_r$) and linear regression model (\emph{lm}).
Figure \ref{fig:relu2} (\emph{Right}) shows RMSE across different methods on train and test data splits. This indicates that our proposed method performs better than \emph{dt}, \emph{rf}, \emph{lm}, \emph{svm}$\_l$ and are comparable to \emph{svm}$\_r$ on the simulated data in terms of RMSE of prediction. The mean and standard deviation of the RMSE for each method are reported in Table \ref{tab:sim1} of Appendix A.

   \begin{figure}[ht!]
     \centering 
\includegraphics[width=1\linewidth]{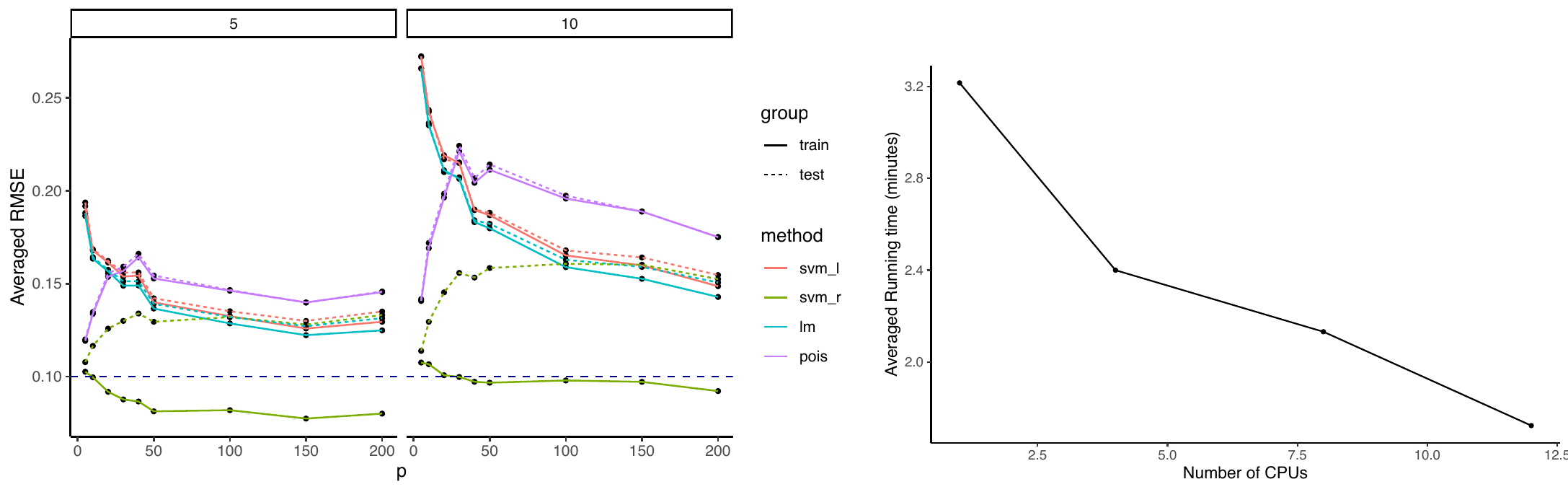}
     \caption{ \emph{Left)} Mean RMSE across different methods on train and test data splits when $|\boldsymbol{P}|=5,~10$ and $p$ varies  in the second simulation study. The solid lines represent the RMSE on the train datasets and the dotted ones indicate the values with respect to the test datasets. \emph{Right)} Running time (in minutes) of our proposed methods with different number of CPUs when $p=20$, $|\boldsymbol{P}|=5$, $L=1000$, and $R=100$. All experiments were run on Digital Alliance Canada's cedar cluster, with memory size set to 8G for each job. }
     \label{fig:sim2_1}
 \end{figure}
 
 In our second simulation study, we compare the performance of our proposed method to existing machine learning methods by varying the number of predictors and the number of hyperplanes. We also include the support vector machines with radial kernels (\emph{svm}$\_r$) for comparison. In this simulation study, each dataset is  simulated by randomly drawing $5,000$ paired coordinates $\boldsymbol{x}=(x_1,x_2,\ldots,x_p)$ from the domain $[-1,-1]^p$.  The response values are given by 
 
 $$y=w_0+w_1\delta(\langle\boldsymbol{x},~\overset{ }{\boldsymbol{n}_1}\rangle-\mu_1)+\ldots+w_{|\boldsymbol{P}|}\delta(\langle\boldsymbol{x},~\overset{ }{\boldsymbol{n}_{|\boldsymbol{P}|}}\rangle-\mu_{|\boldsymbol{P}|})+\epsilon,$$
 where $({\boldsymbol{n}_i},\mu_i)$, $i=1,\ldots, |\boldsymbol{P}|$ are paired normal vectors and distance parameters of hyperplanes sampled from the domain uniformly,  $w_i$ $(i=1, \ldots,|\boldsymbol{P}|)$ is drawn from the standard normal distribution, $w_0$ and the noise $\epsilon$ is sampled from $N(0, 0.1^2)$.  %To simplify the notation, we write $w_0+w_1\delta(\langle\boldsymbol{x},~\overset{ }{\boldsymbol{n}_1}\rangle-\mu_1)+\ldots+w_{|\boldsymbol{P}|}\delta(\langle\boldsymbol{x},~\overset{ }{\boldsymbol{n}_{|\boldsymbol{P}|}}\rangle-\mu_{|\boldsymbol{P}|})$ as $\phi(\boldsymbol{x};\boldsymbol{\theta})$, where $\boldsymbol{\theta}=(\boldsymbol{n}_1,\ldots,\boldsymbol{n}_{|\boldsymbol{P}|},mu_{1},\ldots,mu_{|\boldsymbol{P}|},\sigma^2)$.  
  The number of predictors $p$ and hyperplanes $|\boldsymbol{P}|$ are set to $5$, $10$, $20$, $30$, $40$, $50$, $100$, $200$ and $5,~10$ separately. We repeat the experiment $100$ times for each set of $p$ and $|\boldsymbol{P}|$. Figure \ref{fig:sim2_1}  gives the results of top three methods concerning the RMSE, which are our proposed method, \emph{svm\_l}, and \emph{lm}. The full results are given in Figure \ref{fig:sim2_all} in Appendix. The results indicate that the machine learning methods \emph{svm\_l}, \emph{svm\_r}, \emph{rf}, \emph{dt}, \emph{lm} suffer from overfitting, especially when the number of predictors  $p$ is large. Our proposed method performs consistent on both the train and test datasets when $\boldsymbol{P}$ and $p$ varies. We also conduct the Wilcoxon sign-rank test to compare the performance of our proposed method to \emph{svm\_r}, \emph{svm\_l}, and \emph{lm}. The result is given in Table \ref{tab:sim2} of Appendix and it shows that generally our proposed method performs significant better than the \emph{svm\_l} and \emph{lm} when $p$ is small ($\le 20$).

In our third simulation study, we demonstrate that the performance of our proposed methods when we incorporate decomposition propositions described in Section \ref{sec:DP} into the model. The coordinates $\boldsymbol{x}$ and the response values  $y$ are generated in the same way as the second simulation study with $p=2$, $\boldsymbol{|P|}=40$ and $n=5000$. The decomposition propositions are applied in two ways, we denote it as $decmp1$, $decmp2$ respectively. Firstly, we decompose a large PPP with $\boldsymbol{|P|}=40$ into $4$ independent PPPs with $10$ hyperplane realizations for each, and denote the parameter estimates of each process as $(\hat{w}_0^{i},\hat{w}_1^{i},\hat{\boldsymbol{n}}^{i}_1,\hat{\mu}^{i}_1,\ldots,\hat{w}_{10}^{i},\hat{\boldsymbol{n}}^{i}_{10},\hat{\mu}^{i}_{10}),~i=1,\ldots,4.$ For any given data point $\boldsymbol{x}$, the fitted value of $y$ is given by aggregating the results of $4$ 
independent PPPs in the following manner,
$$\hat{y}=\sum_{i=1}^{4}\left(\frac{\hat{w}_0^{i}}{4}+\sum_{j=1}^{10}\frac{\hat{w}_1^{i}}{4}\delta(\langle\boldsymbol{x},~\overset{ }{\hat{\boldsymbol{n}}^{i}_j}\rangle-\hat{\mu}^{i}_j)\right).$$  
In the second way of decomposition, we allocate the $5,000$ data points to sub-datasets by dividing the coordinate domain into $4$ disjoint sub-domains, denoted as $\mathbb{D}_1$, \ldots,$\mathbb{D}_4$. Here we split the domain evenly into $4$ subdomains along the axis of the $1$st coordinate. For each sub-dataset, we apply the process with $10$ hyperplane realizations, and denote the parameter estimates of each process as $(\hat{w}_0^{i},\hat{w}_1^{i},\hat{\boldsymbol{n}}^{i}_1,\hat{\mu}^{i}_1,\ldots,\hat{w}_{10}^{i},\hat{\boldsymbol{n}}^{i}_{10},\hat{\mu}^{i}_{10}),~i=1,\ldots,4.$ For any given data point $\boldsymbol{x}$, the fitted value of $y$ is given by aggregating the results of $4$ 
independent PPPs in the following manner,
$$\hat{y}=\sum_{i=1}^{4}\boldsymbol{1}_{\mathbb{D}_i}(\boldsymbol{x})\left(\hat{w}_0^{i}+\sum_{j=1}^{10}\hat{w}_1^{i}\delta(\langle\boldsymbol{x},~\overset{ }{\hat{\boldsymbol{n}}^{i}_j}\rangle-\hat{\mu}^{i}_j)\right).$$  We compare the performance of the two decomposition schemes described above to the model when we apply a PPP with $40$ hyperplane realizations to the whole dataset. The latter is referred as \emph{whole}. We repeat the experiment $100$ times, and Figure~\ref{fig:decmps} (\textit{a}) gives the boxplots of RMSE of the two decomposition approaches and the \emph{whole} model. The two decomposition approaches provide similar results compared to the \emph{whole} model.
The mean and standard deviation of the RMSE for each method are reported in Table \ref{tab:sim3} of Appendix A.
{We also evaluate the running time of our decomposition approaches compared to the \emph{whole} model.  Figure~\ref{fig:decmps}(\textit{b}) gives the boxplot of running times ratios for the two decomposition approaches of every single job compared to the \emph{whole} model. The baseline is the running time for every single job of the \emph{whole} model.  As indicated in Figure~\ref{fig:decmps}(\textit{b}), decomposition approaches are more computationally efficient in terms of the running time, and the mean running time ratios are  $13.4$\%,  $4.69$\% for \emph{decomp1} and \emph{decomp2}, separately and the corresponding standard deviations are $4.55$\% and $1.44$\%, separately. Figure~\ref{fig:decmps}(\textit{b}) also shows that \emph{decomp2} is more computationally efficient than \emph{decomp1}, \ie~the decomposition by coordinate domain
is more efficient than the decomposition by the intensity of hyperplane.

 \begin{figure}[ht!]
     \centering
\includegraphics[width=1\linewidth]{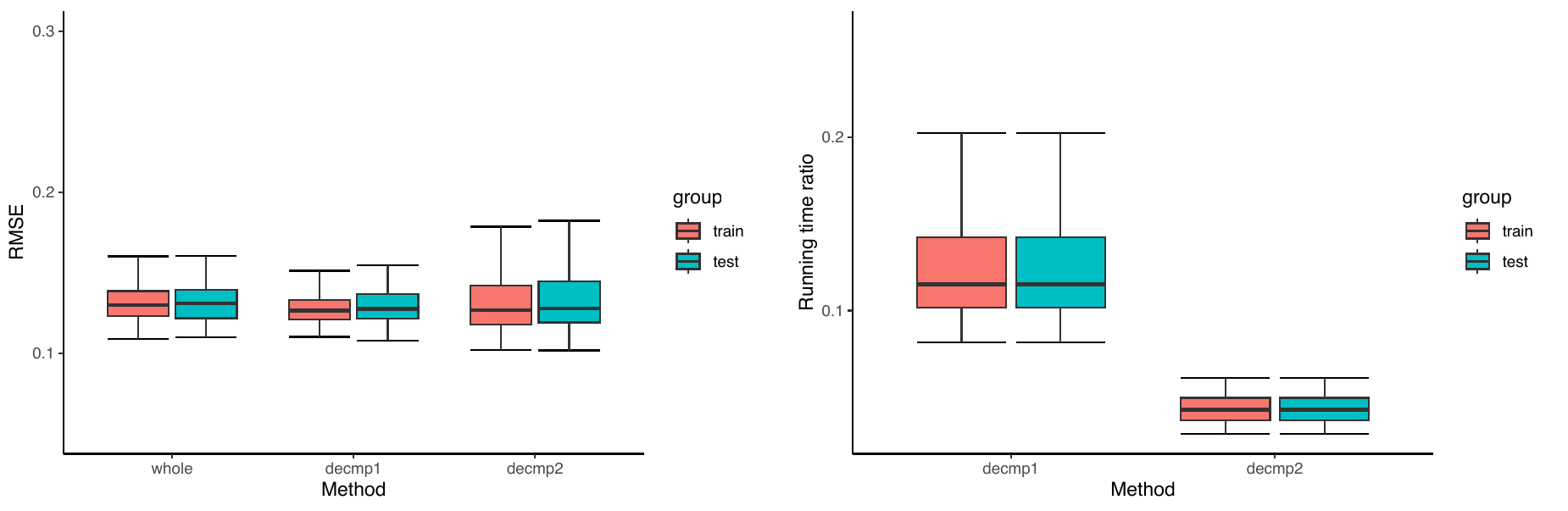}
    \caption{\emph{Left)} The RMSE of the two decomposition approaches compared to the \emph{whole} model on train and test data from $100$ random experiments for the third simulation study. \emph{Right)}  Running time ratios for decomposition approaches of every single job compared to the \emph{whole} model. The baseline is the running time for every single job of the \emph{whole} model. All experiments were run on Digital Alliance Canada's cedar cluster, with memory size set to 8G for each job.  }    
     \label{fig:decmps}
 \end{figure}

In addition, as a by-product of the annealed SMC, we can obtain a set of weighted samples of model parameters, which can be used to assess the uncertainty of predictions (\eg~$95\%$ credible intervals (CIs)).  For non-Bayesian methods,  the bootstrap method \citep{10.1214/aos/1176344552, aslett2021} is often used for constructing confidence intervals (also denoted as CIs).

In our fourth simulation study, we compare the performance of  prediction uncertainty of our proposed method with \emph{nn} and XGBoost (denoted as \emph{xgb}) \citep{10.1145/2939672.2939785} on $10$ random selected datasets (shown in Figure \ref{fig:sim}). 
We set the number of particles to $1000$ and the length of the power series to $100$ for our proposed method. The effective sample size (\emph{the equivalent size of independent samples}) of these $1000$ particles on all these $10$ datasets are approximately $1000$.

  To obtain the $95\%$ CIs for non-Bayesian methods, we set the number of bootstrap samples to $1000$ for the \emph{xgb}, and to $100$ for the \emph{nn} for the sake of easily accessible computational resources.  All these experiments are conducted on a 2.3 GHz Intel Core i9 processor. The results are provided in Figure \ref{fig_rev2}.

  As shown in Figure \ref{fig_rev2} (a), the \emph{pois} has lower RMSE than the \emph{nn}. Although the \emph{xgb} has the smallest RMSE on the train datasets, it shows overfitting on the test datasets (also shown in Figures \ref{fig_rev2} (a, c)). Figure \ref{fig_rev2} (b) indicates that our proposed method was computationally comparable to the \emph{xgb} with $1000$ bootstrap samples. And even with a smaller number ($100$) of bootstrap samples, the running time of the \emph{nn} is over $20$ times than \emph{pois}. Figure \ref{fig_rev2} (c, d) reveal that the coverage rates of $95\%$ CI of the prediction $y$ of the \emph{pois} are around $95\%$ on the $10$ datasets with the smallest length of CIs, while the \emph{xgb} shows larger coverage rates with wider CIs. The \emph{nn} has approximately $100\%$ coverage rates, which could be explained by the large lengths of CIs (\emph{i.e.} large variance). In summary, the \emph{pois} outperforms the \emph{nn} in terms of the measure of uncertainty concerning metrics RMSE, running times, and the coverage of predictions. On the test datasets, the \emph{pois} slightly outperforms the \emph{xgb} in terms of the model accuracy (\emph{e.g}, lower RMSE, more accurate  CIs). Table \ref{tab_rev2} provides the mean and standard deviation of methods \emph{pois}, \emph{xgb} and \emph{nn} concerning uncertainty.

\begin{figure}[h]
    \centering
    \includegraphics[width=1\linewidth]{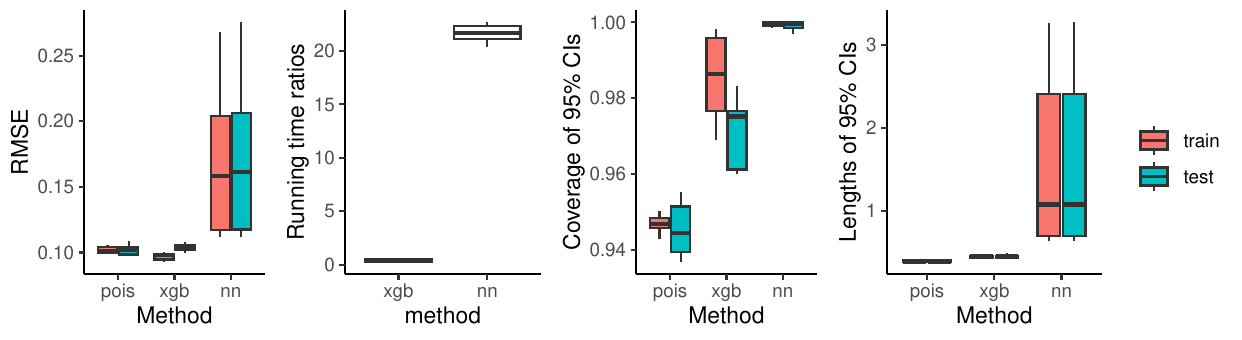}
    \caption{Comparison across our proposed method (\textit{pois}), xgboost (\textit{xgb}) and a two layer ReLU NN (\textit{nn}) in the fourth simulation. (a) RMSE on the train datasets and test datasets. (b) Running time ratios
    for competitors over our proposed method.  (c - d) Coverage rates and lengths  of the $95\%$ CI of the predictions for different methods.}
    \label{fig_rev2}
\end{figure}

\begin{table}[h]
    \centering
    \begin{tabular}{l|c c|c c}
    \hline
       metric $\&$ method   &  Train & & Test & \\
                & mean & sd & mean & sd\\   
       \hline 
        RMSE &&&   \\
\emph{pois} & 0.102  &0.0028  & 0.102 &0.0040\\
\emph{xgb} &0.097 &0.0027 & 0.104 &0.0030\\
 \emph{nn} &0.167 &0.0561 &0.170 &0.0579\\   
        \hline 
        Running time (seconds) & &&  \\
 \emph{pois} & 45    & 1.15  &  &  \\
\emph{xgb} & 18.1   &  5.03 &    &  \\
 \emph{nn} & 976   & 19.5  &  &  \\    
                \hline 
        Coverage rate of $95\%$ CIs  & &&  \\
         \emph{pois} &0.947 &0.00291  &0.945 & 0.00697 \\
\emph{xgb} & 0.985 &0.0114  & 0.971 &0.00901 \\
 \emph{nn} &1.00 & 0.000577 &0.999 &0.00107 \\   
               \hline 
        Lengths of $95\%$ CIs & && \\
         \emph{pois} & 0.394 & 0.0104 & 0.394& 0.0103\\
\emph{xgb} &  0.448 & 0.0188 & 0.450 & 0.0203\\
 \emph{nn} &  1.57 & 1.08  & 1.57 & 1.08 \\             
               \hline 
    \end{tabular}
    \caption{Mean and standard deviation of methods \emph{pois}, \emph{xgb} and \emph{nn} concerning metrics RMSE, running times (in seconds), coverage rates and length of the $95\%$ CI of the predictions.}
    \label{tab_rev2}
\end{table}

%To simplify the notation, we write $w_0+w_1\delta(\langle\boldsymbol{x},~\overset{ }{\boldsymbol{n}_1}\rangle-\mu_1)+\ldots+w_{|\boldsymbol{P}|}\delta(\langle\boldsymbol{x},~\overset{ }{\boldsymbol{n}_{|\boldsymbol{P}|}}\rangle-\mu_{|\boldsymbol{P}|})$ as $\phi(\boldsymbol{x};\boldsymbol{\theta})$, where $\boldsymbol{\theta}=(\boldsymbol{n}_1,\ldots,\boldsymbol{n}_{|\boldsymbol{P}|},mu_{1},\ldots,mu_{|\boldsymbol{P}|},\sigma^2)$.  
 
\section{Application}
\label{sec: app}
 \begin{figure}[ht!]
     \centering
 \includegraphics[width=1\linewidth]{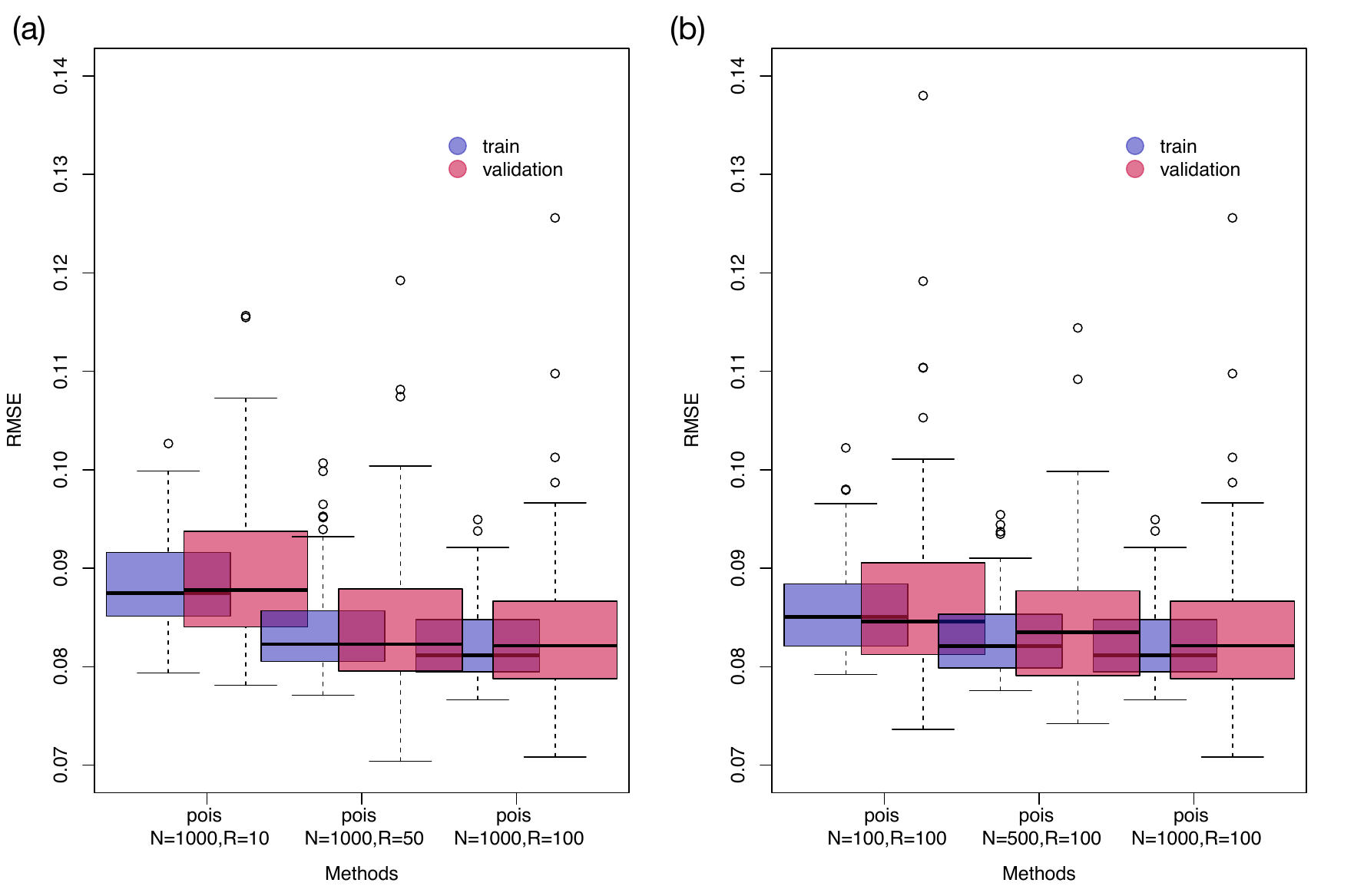}
     \caption{The RMSE of our proposed method on $100$ train and test splits of abalone data. (a-b) The box plots of the RMSE of our proposed method when the number of particles and the length of the power series vary. }
     \label{fig:abalone_ab}
 \end{figure}

 \begin{figure}[ht!]
     \centering
 \includegraphics[width=1\linewidth]{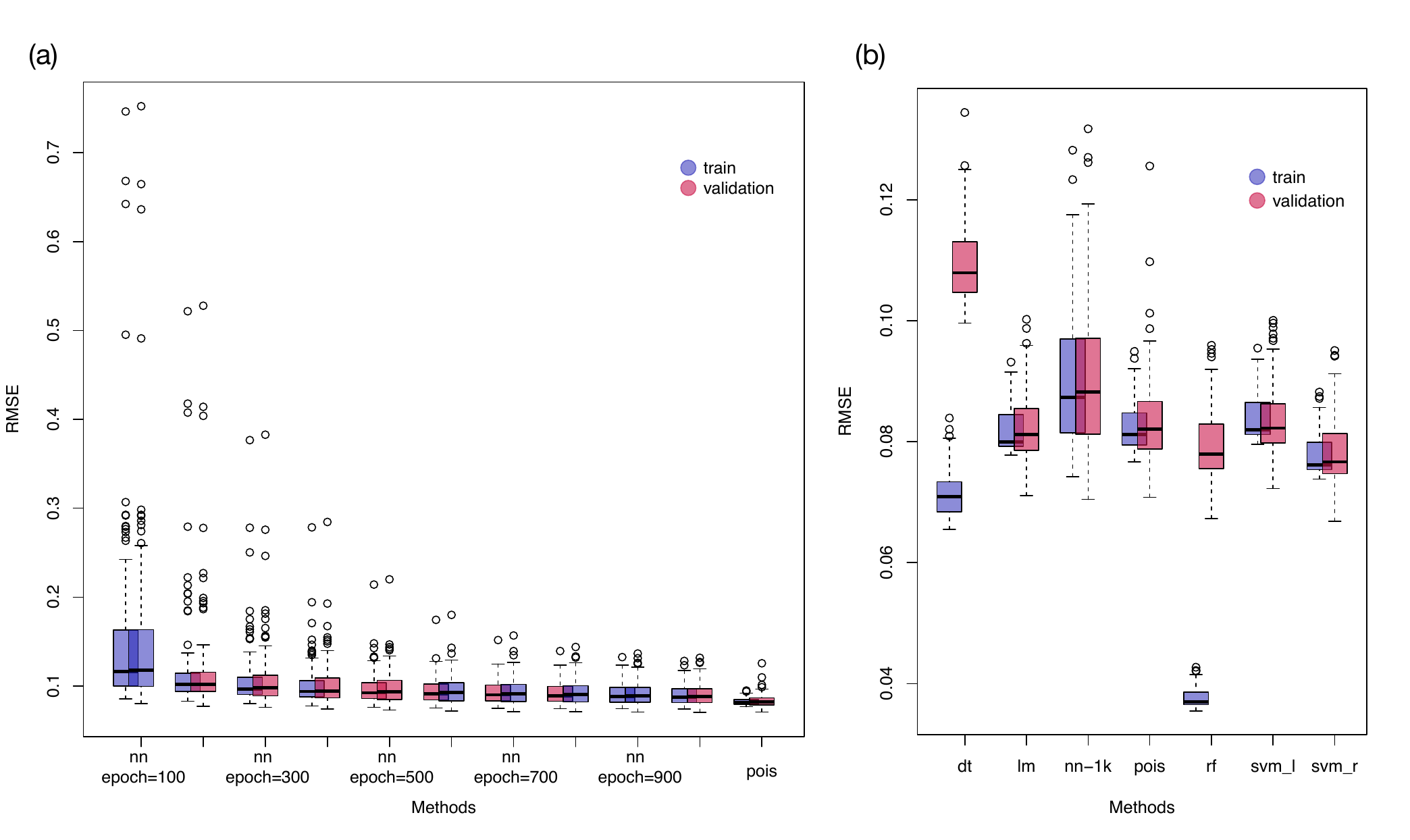}
     \caption{The RMSE across different methods on $100$ train and test splits of abalone data. (a) The box plots of the RMSE of the two-layer ReLU NN with different number of epochs compared to our proposed method with $L=1000$, $R=100$. (b)  The RMSE across different methods. The number of epochs of the two-layer ReLU NN is fixed to $1000$, and $L$, $R$ of our proposed method is fixed to $1000$ and  $100$, respectively.     }
     \label{fig:abalone_cd}
 \end{figure}
In this section, we use real data sets to evaluate the performance of our proposed method and the two-layer ReLU NN (\emph{nn}) implemented in {\tt keras.Sequential()} of the package Tensorflow (version 2.9.1) with Adam optimization \citep{tensorflow2015,kingma2014adam}. 
The comparison is based on two real datasets from UCI Machine Learning Repository: the red wine dataset \citep{misc_wine_quality_186} and the abalone dataset \citep{misc_abalone_1}. The red wine data contains 11 features (fixed acidity,  volatile acidity, citric acid, residual sugar, chlorides, free sulfur dioxide, total sulfur dioxide, density, pH values, sulphates and  alcohol) and the response is the quality score that represents the quality of wine with 1599 samples. The abalone dataset contains physical measures of 4177 Tasmanian abalones, and our goal is to predict the age (the rings) of abalone given 7 continuous physical measurements, including length (mm), diameter (mm), height (mm), whole weight (grams), meat weight (grams), gut weight (grams) and shell weight (grams). For each dataset, we randomly generate $100$ train/test splits, with each train and test data containing 75\% and 25\% of the samples. Figures \ref{fig:abalone_ab}-\ref{fig:abalone_cd} and Figures \ref{fig:redwine_ab}-\ref{fig:redwine_cd} represent the RMSE of each method over the 100 splits for the two data sets respectively.  The  number of hyperplanes of our proposed methods is fixed to $5$ in these experiments. The mean and standard deviation of the RMSE for each method are reported in Table \ref{tab: abalone} and Table  \ref{tab: wine} of Appendix A. Panels (a) and (b) of both figures \ref{fig:abalone_ab},~\ref{fig:redwine_ab} indicate that increasing the number of particles or the length of power series improve the prediction performance of our proposed method (\emph{pois}). Panel (a) of both figures \ref{fig:abalone_cd},~\ref{fig:redwine_cd} demonstrate that our model outperform the ReLU NN with various number of epochs. For the abalone data, our model performs better than \emph{dt} and \emph{nn}, and are comparable to \emph{lm}, \emph{svm}$\_r$ and \emph{svm}$\_l$, 
as shown in panel (b) of Figure \ref{fig:abalone_cd}. 
For the red wine data, our model performs better than \emph{dt}, and are comparable to  \emph{nn}, \emph{lm} and \emph{svm}$\_l$, 
as shown in panel (b) of Figure \ref{fig:redwine_cd}.

 \begin{figure}[ht!]
     \centering
 \includegraphics[width=0.95\linewidth]{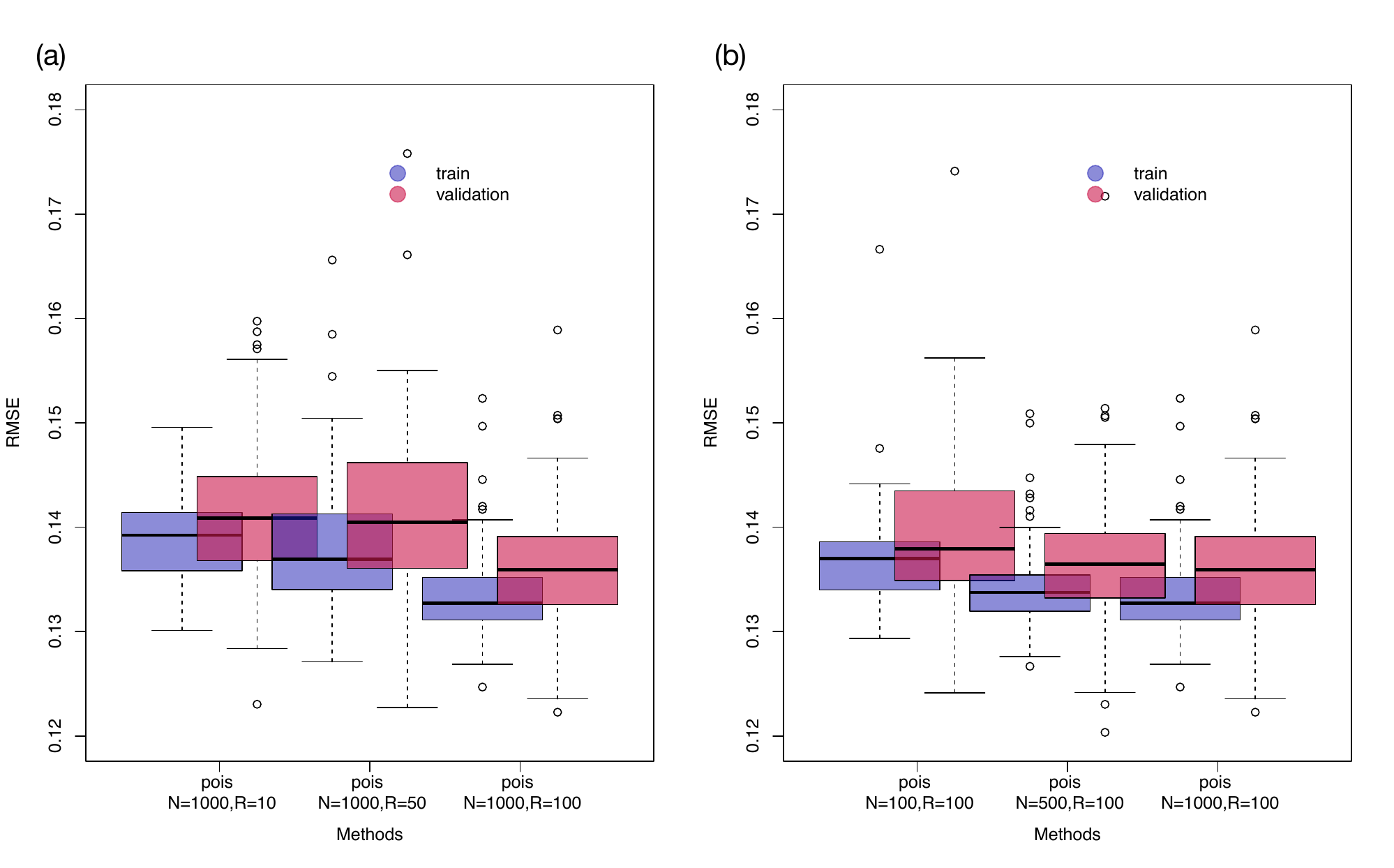}
     \caption{The RMSE of our proposed method on $100$ train and test splits of red wine data. (a-b) The box plots of the RMSE of our proposed method when the number of particles and the length of the power series vary.  }
     \label{fig:redwine_ab}
 \end{figure}

 \begin{figure}[ht!]
     \centering
 \includegraphics[width=0.95\linewidth]{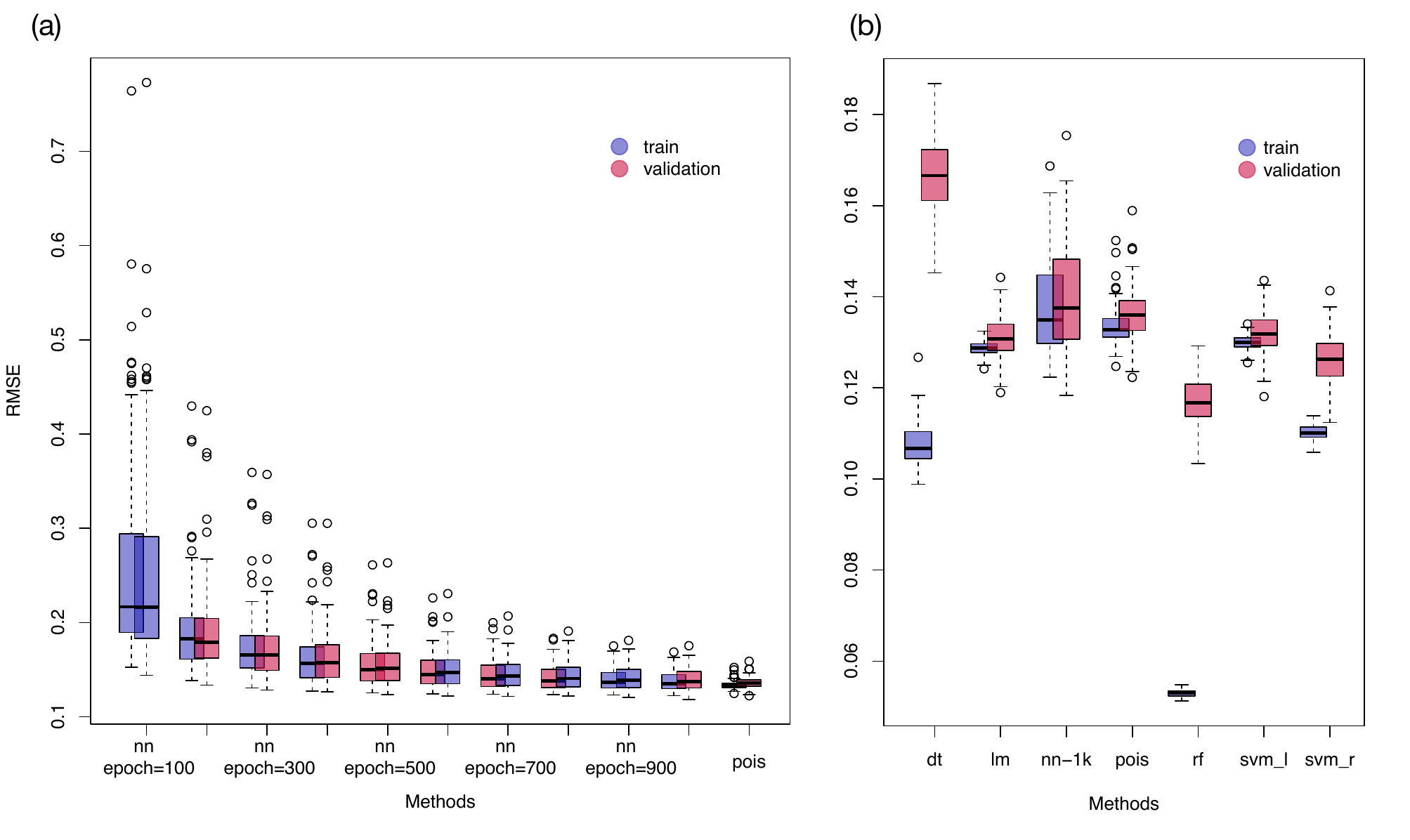}
     \caption{The RMSE across different methods on $100$ train and test splits of red wine data.  (a) The box plots of the RMSE of the two-layer ReLU NN with different numbers of epochs compared to our proposed method with $L=1000$, $R=100$. (b)  The RMSE across different methods. The number of epochs of the two-layer ReLU NN is fixed to $1000$   and $L$, $R$ of our proposed method is fixed to $1000$ and  $100$, respectively. 
     }
     \label{fig:redwine_cd}
 \end{figure}
 
 \section{Discussion}
 \label{sec:Dis}
We provide a framework for viewing a two-layer ReLU neural network as Poisson hyperplane processes with a Gaussian prior. 
We propose three decomposition propositions to show that a two-layer NN constructed by PHP can be adapted to large scale problems. Note that weight tying is a technique similar to our method, and often used to reduce the size of neural translation models
to less than the original size without harming the performance \citep{press2017usingoutputembeddingimprove}. The improvement is achieved by sharing the weights of input and output embeddings.
The idea of the decomposition scheme indicated by \textbf{Proposition}~\ref{prop: 1} is to use multiple simpler models (with fewer hyperplanes) to model the data, and aggregate these models to make the final prediction. The simpler models are independent, hence they do not share parameters. 
For the decomposition scheme indicated by \textbf{Proposition}~\ref{prop: 3}, we split the domain and use different PHPs to model each domain, no parameters are shared across PHPs.
In this way, the two decomposition schemes are not identical to the weight tying technique.

The model inference is performed via a sequential Monte Carlo algorithm. As demonstrated by our simulation study and real applications, 
the proposed method can improve performance over the two-layer ReLU NN and other classical 
machine learning methods. 
 In our model, the number of hyperplanes is  prespecified (\emph{i.e.} according to prior knowledge or tuning) before running the experiments. One future extension is to treat 
        the number of hyperplanes as an unknown parameter, and add a prior distribution on  
        it. Reversible jump MCMC will be designed to draw the inference, the number of hyperplanes will be determined by the likelihood model and the prior distribution. 
        Another line of future work includes proposing PHP for online fashion data.

\section*{Funding}
%SG was supported by the Shanghai Science and Technology Program (No. 21010502500), the National Natural Science Funds of China (12401383), the startup fund of ShanghaiTech University,  SW was supported by the National Natural Science Funds of China (No. 12101333, No.62176068), the Natural Science Funds of Tianjin (No. 21JCQNJC00050), LTE was supported by a Natural Sciences and Engineering Research Council of Canada (NSERC) award DGECR/00118-2019 and a Michael Smith Health Research BC Scholar Award.
This work was supported by the National Natural Science Foundation of China (12401383 and
12101333), the startup fund of ShanghaiTech University, the Natural Sciences and Engineering Research Council of Canada (NSERC) award DGECR/00118-2019 and a Michael Smith Health Research BC Scholar Award. This research was enabled in part by support provided by Digital Research Alliance of Canada.

\section*{Conflict of interest}
The authors declare that they have no conflict of interest.

\bibliographystyle{spbasic}      % basic style, author-year citations
\bibliography{document}
%\begin{thebibliography}{}
%%
%% and use \bibitem to create references. Consult the Instructions
%% for authors for reference list style.
%\end{thebibliography}

\appendix

\section{Additional numerical results}
  \begin{figure}[ht!]
     \centering
\includegraphics[width=0.85\linewidth]{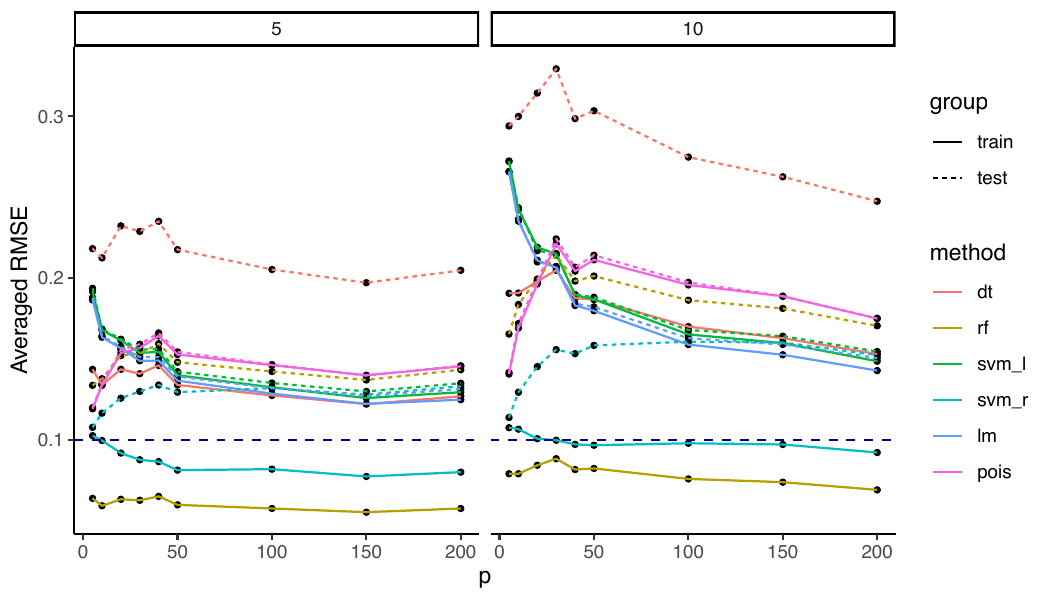}
     \caption{Mean RMSE across different methods on train and test data splits when $|\boldsymbol{P}|=5,~10$ and $p$ varies  in the second simulation study. The solid lines represent the RMSE on the train datasets and the dotted ones indicate the values with respect to the test datasets.}
     \label{fig:sim2_all}
 \end{figure}

\begin{table}[ht]
\centering
\caption{Mean and standard error of the RMSE of simulation 1 in Section \ref{sec: sim}.}
\label{tab:sim1}
\begin{tabular}{lrrrr}
 \hline
 %\multicolumn{2}{| c |}{Begin of Table}\\
  \hline
&&   Train  & Test& \\
  \hline
  method & mean & sd & mean & sd\\ 
  \hline
 \emph{pois} - L=1000 R=10  & 0.104 & 0.007  &  0.104 & 0.007 \\  
 \emph{pois} - L=1000 R=50  & 0.102 & 0.005  &  0.102& 0.005 \\ 
 \emph{pois} -  L=1000 R=100  & 0.101 & 0.002 & 0.101 & 0.002 \\ 
 \emph{pois} -  L=2000 R=100  & 0.102 & 0.01  & 0.102 & 0.01 \\ 
  \emph{dt}  & 0.0913 & 0.005  & 0.14 & 0.01 \\ 
   \emph{rf}  & 0.0518 & 0.001  & 0.107 & 0.002 \\ 
  \emph{lm}  & 0.154 & 0.074  & 0.154 & 0.073 \\ 
 \emph{svm\_l}  & 0.157 & 0.08 &  0.157 & 0.08 \\ 
  \emph{svm\_r}  & 0.101 & 0.002 &  0.101 & 0.002 \\ 
   \hline
\end{tabular}
\end{table}

\begin{table}[ht]
\centering
\caption{P-values of the Wilcoxon sign-rank test of the RMSE of $svm_l$ ($svm_r$, $lm$) v.s. $pois$ on the testing data of simulation 2. The alternative hypothesis is that our proposed method has smaller RMSE than the $svm_l$ ($svm_r$, $lm$). Symbols *,**,*** indicate that the null hypothesis is rejected at significant level $0.001$, $0.05$,$0.01$ respectively. }
\label{tab:sim2}
{\small
\begin{tabular}{rrlll}
  \hline
  $|\boldsymbol{P}|$ &p & $svm_l$ v.s. $pois$ & $svm_r$ v.s. $pois$ & $lm$ v.s. $pois$\\
  \hline
   5 &  5 & *** & 1 & *** \\ 
   5 & 10 & *** & 1 & *** \\ 
   5 & 20 & *** & 1 & * \\ 
   5 & 30 & 0.969 & 1 & 1 \\ 
   5 & 40 & 1 & 1 & 1 \\ 
   5 & 50 & 1 & 0.997 & 1 \\ 
   5 &100 & 0.953 & 0.927 & 0.991 \\ 
   5 &150 & 0.301 & 0.464 & 0.606 \\ 
   5 &200 & 0.044 & 0.272 & 0.230 \\ 
  10 &  5 & *** & 1 & *** \\ 
  10 & 10 & *** & 1 & *** \\ 
  10 & 20 & *** & 1 & *** \\ 
  10 & 30 & 1 & 1 & 1 \\ 
  10 & 40 & 1 & 1 & 1 \\ 
  10 & 50 & 1 & 1 & 1 \\ 
  10 &100 & 1 & 1 & 1 \\ 
  10 &150 & 1 & 1 & 1 \\ 
  10 &200 & 0.999 & 1 & 1 \\  
   \hline
\end{tabular}
}
\end{table}

\begin{table}[ht]
\centering
\caption{Mean and standard error of the RMSE of simulation 3 in Section \ref{sec: sim}.}
\label{tab:sim3}
\begin{tabular}{lrrrr}
 \hline
 %\multicolumn{2}{| c |}{Begin of Table}\\
  \hline
&&   Train  & Test& \\
  \hline
  method & mean & sd & mean & sd\\ 
  \hline
 \emph{whole}   & 0.131 & 0.0130    &  0.133 &  0.0137 \\ 
  \emph{decmp1} & 0.128 &  0.0107  &  0.129 &  0.0113\\
   \emph{decmp2}  &  0.133 &  0.0249  & 0.135 &  0.0250 \\
   \hline
\end{tabular}
\end{table}

%\begin{table}[ht]
%\centering
%\caption{Mean and standard error of running time ratios of single jobs of the decomposition approaches compared to the whole model of simulation 3 in Section \ref{sec: sim}.}
%\label{tab:sim3_b}
%\begin{tabular}{lrrrr}
% \hline
% %\multicolumn{2}{| c |}{Begin of Table}\\
%  \hline
%  method & mean & sd \\
%  \hline
%  \emph{decmp1} & 0.134 &  0.0455  \\
%   \emph{decmp2}  &  0.0469 &  0.0144   \\
%   \hline
%\end{tabular}
%\end{table}

% latex table generated in R 4.3.0 by xtable 1.8-4 package
% Wed Nov  1 16:24:31 2023
\begin{table}[ht]
\centering
\caption{The RMSE mean and standard deviation (SD) of the 100 train and test splits of the abalone data of each method.}
\label{tab: abalone}
\begin{tabular}{llrrrr}
  \hline
  \hline
&&   Train & & Test& \\
  \hline
  method  & mean & sd & mean & sd\\ 
 %  Method & Mean(train) & SD(train)  & Mean(test) & SD(test)\\ 
 \hline
\emph{pois} - L=500, R= 100    & 0.08 & 0.00 & 0.08 & 0.01 \\ 
\emph{pois} - L=1000, R= 10    & 0.09 & 0.00 & 0.09 & 0.01 \\ 
\emph{pois} - L=1000, R= 50    & 0.08 & 0.00 & 0.08 & 0.01 \\ 
\emph{pois} - L=1000, R= 100   & 0.08 & 0.00 & 0.08 & 0.01 \\ 
   \emph{dt} & 0.07 & 0.00 & 0.11 & 0.01 \\ 
   \emph{lm} & 0.08 & 0.00 & 0.08 & 0.01 \\
   \emph{rf} & 0.04 & 0.00 & 0.08 & 0.01 \\ 
   \emph{svm\_l} & 0.08 & 0.00   & 0.08 & 0.01 \\ 
   \emph{svm\_r} & 0.08 & 0.00  & 0.08 & 0.01 \\ 
   \emph{nn} - \#epoc=100 & 0.16 & 0.11 & 0.16 & 0.11 \\ 
   \emph{nn} - \#epoc=200 & 0.12 & 0.07 & 0.12 & 0.07\\ 
   \emph{nn} - \#epoc=300 & 0.11 & 0.04 & 0.11 & 0.04\\ 
   \emph{nn} - \#epoc=400 & 0.10 & 0.03 & 0.10 & 0.03 \\ 
   \emph{nn} - \#epoc=500 & 0.10 & 0.02 & 0.10 & 0.02\\ 
   \emph{nn} - \#epoc=600 & 0.10 & 0.02 & 0.10 & 0.02\\ 
   \emph{nn} - \#epoc=700 & 0.09 & 0.01 & 0.09 & 0.02\\ 
   \emph{nn} - \#epoc=800 & 0.09 & 0.01 & 0.09 & 0.01 \\ 
   \emph{nn} - \#epoc=900 & 0.09 & 0.01 & 0.09 & 0.01 \\ 
   \emph{nn} - \#epoc=1000 & 0.09 & 0.01 & 0.09 & 0.01 \\ 
   \hline
\end{tabular}
\end{table}

\begin{table}[ht]
\centering
\caption{The RMSE mean and standard deviation (SD) of the 100 train and test splits of the red wine data of each method.}
\label{tab: wine}
\begin{tabular}{llrrrr}
  \hline
    \hline
&&   Train & & Test& \\
  \hline
  method  & mean & sd & mean & sd\\
   %Method & Mean(train) & SD(train)  & Mean(test) & SD(test)\\ 
 \hline
 \emph{pois} - L=100, R= 100 & 0.14 & 0.00   & 0.14 & 0.01 \\ 
\emph{pois} - L=500, R= 100 & 0.13 & 0.00 & 0.14 & 0.01\\ 
\emph{pois} - L=1000, R= 10 & 0.14 & 0.00  & 0.14 & 0.01 \\ 
\emph{pois} - L=1000, R= 50 & 0.14 & 0.01 & 0.14 & 0.01 \\ 
\emph{pois} - L=1000, R= 100 & 0.13 & 0.00  & 0.14 & 0.01  \\ 
%\emph{pois} - N=100, R= 100 100\_100.t & 0.09 & 0.00  & 0.09 & 0.01 \\ 
    \emph{dt}  & 0.11 & 0.00& 0.17 & 0.01 \\ 
   \emph{lm}  & 0.13 & 0.00  & 0.13 & 0.00 \\ 
   \emph{rf}  & 0.05 & 0.00 & 0.12 & 0.01 \\ 
   \emph{svm\_l}  & 0.13 & 0.00  & 0.13 & 0.01 \\ 
  \emph{svm\_r}  & 0.11 & 0.00  & 0.13 & 0.01 \\ 
 \emph{nn} - \#epoc=100 & 0.26 & 0.11 & 0.26 & 0.11 \\ 
 \emph{nn} - \#epoc=200 & 0.19 & 0.05 & 0.19 & 0.05 \\ 
 \emph{nn} - \#epoc=300 & 0.17 & 0.04 & 0.17 & 0.04 \\ 
 \emph{nn} - \#epoc=400 & 0.16 & 0.03 & 0.16 & 0.03 \\ 
 \emph{nn} - \#epoc=500 & 0.16 & 0.02 & 0.16 & 0.02 \\ 
 \emph{nn} - \#epoc=600 & 0.15 & 0.02  & 0.15 & 0.02 \\ 
 \emph{nn} - \#epoc=700 & 0.15 & 0.02 & 0.15 & 0.02 \\ 
 \emph{nn} - \#epoc=800 & 0.14 & 0.01 & 0.14 & 0.01 \\ 
 \emph{nn} - \#epoc=900 & 0.14 & 0.01 & 0.14 & 0.01 \\ 
 \emph{nn} - \#epoc=1000 & 0.14 & 0.01  & 0.14 & 0.01 \\ 
   \hline
\end{tabular}
\end{table}

\end{document}